\documentclass{article}

\PassOptionsToPackage{numbers,sort&compress}{natbib}

\usepackage[preprint]{neurips_2025}
\makeatletter
\@submissionfalse
\makeatother

\setcitestyle{comma}

\usepackage{authblk}
\usepackage[utf8]{inputenc} 
\usepackage[T1]{fontenc}    
\usepackage{hyperref}       
\usepackage{url}            
\usepackage{booktabs}       
\usepackage{amsfonts}       
\usepackage{xcolor}         
\usepackage{nicefrac}       
\usepackage{microtype}      
\usepackage{xcolor}         
\usepackage{float}
\usepackage{graphicx}
\usepackage{multirow} 
\usepackage{longtable}
\usepackage{mathtools}

\usepackage{setspace} 
\usepackage{array}   

\usepackage{caption}
\usepackage{makecell}

\title{Decoupling Knowledge and Reasoning in LLMs: \\An Exploration Using Cognitive Dual-System Theory}

\begin{document}

\author{}

\maketitle

\vspace{-5em}  
\begin{center}
  \begin{minipage}[t]{0.45\textwidth}
    \centering
    \textbf{Mutian Yang}\\
    Tsinghua University\\
    Beijing, China\\
    \texttt{yangmutian@mail.tsinghua.edu.cn}
  \end{minipage}
  \hfill
  \begin{minipage}[t]{0.45\textwidth}
    \centering
    \textbf{Jiandong Gao}$^\dag$\\
    Tsinghua University\\
    Beijing, China\\
    \texttt{jdgao@tsinghua.edu.cn}
  \end{minipage}

  \vspace{2ex}  

  \begin{minipage}[t]{0.6\textwidth}
    \centering
    \textbf{Ji Wu}$^\dag$\\
    Tsinghua University\\
    Beijing, China\\
    \texttt{wuji\_ee@tsinghua.edu.cn}
  \end{minipage}
\end{center}

\vspace{1em}  

\begin{abstract}
While large language models (LLMs) leverage both knowledge and reasoning during inference, the capacity to distinguish between them plays a pivotal role in model analysis, interpretability, and development. Inspired by dual-system cognitive theory, we propose a cognition attribution framework to decouple the contribution of knowledge and reasoning. In particular, the cognition of LLMs is decomposed into two distinct yet complementary phases: knowledge retrieval (Phase 1) and reasoning adjustment (Phase 2). To separate these phases, LLMs are prompted to generate answers under two different cognitive modes, fast thinking and slow thinking, respectively. The performance under different cognitive modes is analyzed to quantify the contribution of knowledge and reasoning. This architecture is employed to 15 LLMs across 3 datasets. Results reveal: (1) reasoning adjustment is domain-specific, benefiting reasoning-intensive domains (e.g., mathematics, physics, and chemistry) and potentially imparing knowledge-intensive domains. (2) Parameter scaling improves both knowledge and reasoning, with knowledge improvements being more pronounced. Additionally, parameter scaling make LLMs reasoning significantly more prudent, while moderately more intelligent. (3) Knowledge primarily resides in lower network layers, while reasoning operates in higher layers. Our framework not only helps understand LLMs from a "decoupling" perspective, but also provides new insights into existing research, including scaling laws, hierarchical knowledge editing, and limitations of small-model reasoning.
\end{abstract}

\section{Introduction}
\label{Introduction}

LLMs have garnered significant research attention due to remarkable capabilities \cite{achiam2023gpt,yang2024qwen2,touvron2023llama}. Building upon the foundation of scaling laws \cite{kaplan2020scaling}, various methodologies (including pretraining \cite{devlin2019bert}, continual pretraining \cite{ke2023continual,yang2024zhongjing,xie2024efficient}, and instruction tuning \cite{hu2022lora,li2021prefix,liu2021p}) 
are developed to inject knowledge into LLMs. However, their reasoning capability remains disputed, as they often struggle with complex reasoning tasks \cite{mirzadeh2024gsm,huang2025math}.

To address this challenge, Chain-of-Thought (CoT) is proposed, enabling LLMs to mimic human-like progressive reasoning by generating intermediate reasoning steps \cite{wei2022chain}. However, early approaches to CoT generation typically relied on domain-specific prompt engineering, lacking the capability to automatically produce universally applicable reasoning chains across diverse domains.

The emergence of reasoning LLMs, such as OpenAI o1, enables automatic generation of universal CoT through distillation and reinforcement learning \cite{openai2023learningtoreason}. Although the specifics of o1 remain undisclosed, extensive replication efforts have successfully produced LLMs with powerful reasoning capability \cite{qin2024o1,huang2024o1,huang2025o1}. The breakthrough demonstrates that LLMs possess not only extensive knowledge but also advanced reasoning abilities.

In this context, it is scientifically important to distinguish between the contribution of knowledge and reasoning, as this is crucial for understanding the inference behaviours of LLMs. However, the joint employment of knowledge and reasoning during inference make it hard to discern their contribution.

For this purpose, we propose a cognition attribution framework based on dual-system cognitive theory, which decomposes LLMs inference into two distinct but complementary phases: (1) knowledge retrieval (Phase 1), where LLMs rapidly generate initial responses by accessing learned information, and (2) reasoning adjustment (Phase 2), where they refines the initial responses through CoT generation. To separate the two cognitive phases, LLMs are prompted to generate answers under two distinct cognitive modes: fast thinking and slow thinking. During fast thinking, LLMs experience Phase 1, while during slow thinking, LLMs rely on both Phase 1 and Phase 2. The difference between cognitive modes is analyzed to decouple knowledge and reasoning. Our main findings include:

$\bullet$ The contribution of reasoning adjustment varies across domains. It plays more crucial roles in some reasnong-intensive domains (such as mathematics, physics, chemistry) than the others.

$\bullet$ Parameter scaling enhances both knowledge and reasoning, with knowledge being the dominant factor. Additionally, parameter scaling makes the reasoning significantly more "prudent" in all domains and moderately more "intelligent" in some specific domains.

$\bullet$ Knowledge retrieval primarily occurs in lower network layers, while reasoning adjustments are localized in higher layers, suggesting a functional separation in cognition.

In conclusion, our study presents a cognition attribution architecture that decouples knowledge and reasoning in LLMs. This framework not only offers a novel perspective on the cognitive characteristics of LLMs, but also provides new insights into related areas of research, including scaling laws \cite{kaplan2020scaling}, hierarchical knowledge editing \cite{zhang2024co,meng2022locating,meng2022mass}, and limitations of small-model reasoning \cite{li2025small}. Our datasets and code will be publicly released.

\section{From Dual-System Cognitive Theory to Cognition Attribution Architecture}
\label{From Cognition Dual-System Theory to Cognition Attribution Architecture}

\begin{figure}[!h]
    \centering
    \includegraphics[width=1\linewidth]{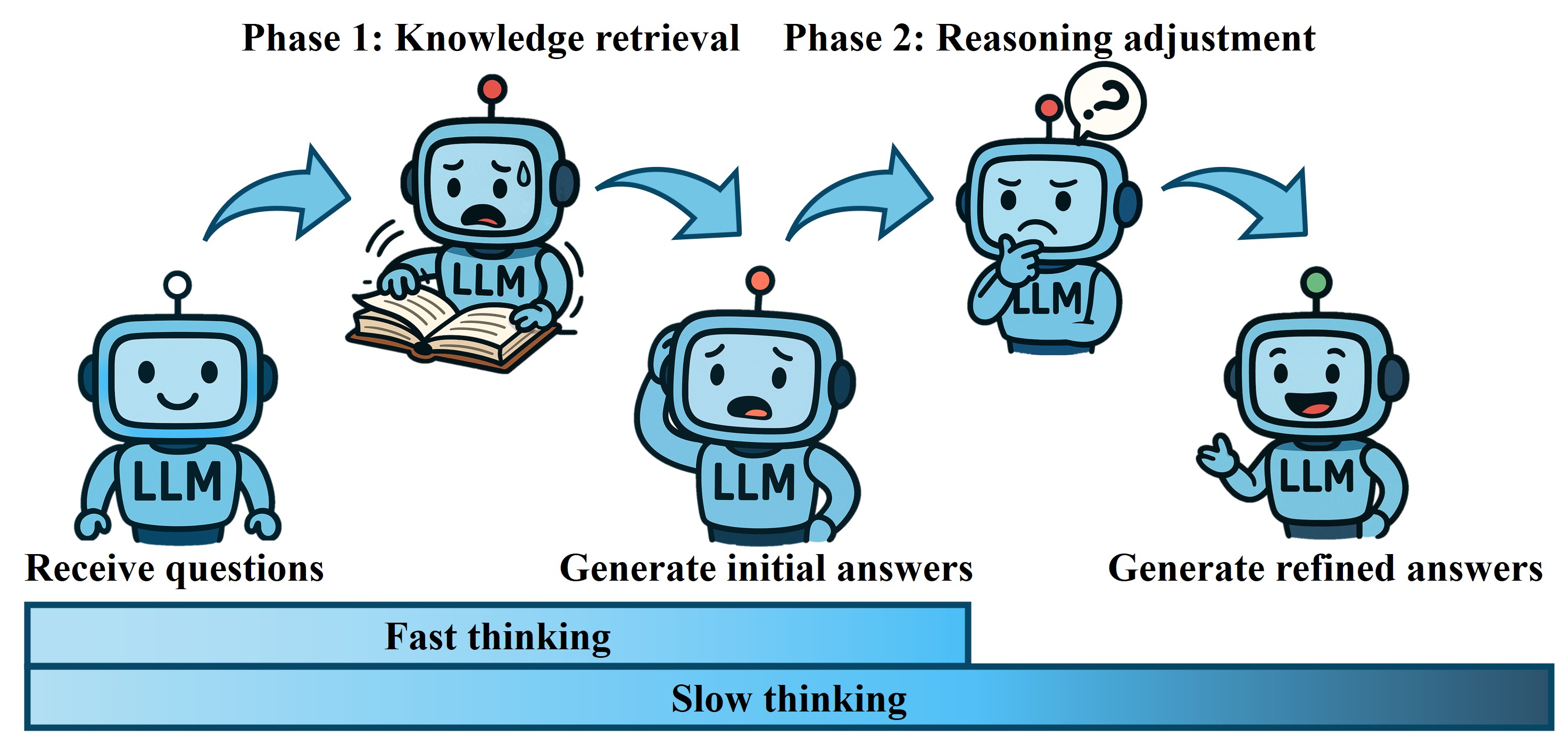}
    \caption{The schema of the cognition attribution architecture, supposing the cognitive process of LLMs involves two primary phases: knowledge retrieval (Phase 1) and reasoning adjustment (Phase 2). In the fast thinking mode, LLMs rely solely on knowledge retrieval. In contrast, in the slow thinking mode, they leverage both knowledge retrieval and reasoning adjustment.}
    \label{fig:model}
\end{figure}

LLMs perform inference through the integrated application of knowledge and reasoning, making it challenging to isolate and evaluate their respective contributions. To address this issue, we propose a cognition attribution architecture based on dual-system cognitive theory, as shown in Figure \ref{fig:model}. This architecture decouples knowledge and reasoning in four steps.

\textbf{Step 1: Define the respective roles of knowledge and reasoning.}

According to dual-system cognitive theory \cite{kahneman2011thinking}, humans exhibit two distinct yet complementary cognitive phases: knowledge retrieval (Phase 1) and reasoning adjustment (Phase 2). This study maps these concepts onto LLMs, hypothesizing LLMs also generate answers in two phases. In Phase 1, LLMs retrieves memorized knowledge purely based on the input. In Phase 2, LLMs applied CoT reasoning to adjust these initial retrievals.

\textbf{Step 2: Separate knowledge retrieval and reasoning adjustment.}

To separate these phases, we further introduce the concept of cognitive modes in dual-system cognitive theory - fast thinking and slow thinking. LLMs are prompted to generate responses under each modes separately (the detailed methods is in Appendix \ref{prompt}).

During fast thinking, LLMs generates responses $y_{fast}$ directly based on the input question $x$ without any additionally reasoning (an example in Appendix \ref{fast thinking exmaple}). In this mode, LLMs rely solely on Phase 1, with its performance depends entirely on the knowledge retrieval capability, denoted as $C_{\text{KR}}$. The process is formally defined as:

\begin{equation}
y_{\text{fast}} = \mathop{\arg\max} P(y \mid x=\text{Question}; C_{\text{knowledge}})
\end{equation}

During slow thinking, LLMs first produces initial answers $y_{fast}$ similar to fast thinking, and then adjusts initial answers through CoT reasoning to get refined answers $y_{slow}$ (an example in Appendix \ref{slow thinking exmaple}). In this mode, LLMs experience both Phase 1 and Phase 2. The performance relies on both knowledge retrieval capability $C_{\text{KR}}$ and reasoning adjustment capability $C_{\text{RA}}$, defined as:

\begin{equation}
y_{\text{slow}} = \mathop{\arg\max} P(y \mid y_{\text{fast}}, x=\text{Question}; C_{\text{knowledge}}, C_{\text{reasoning}})
\end{equation}

\textbf{Step 3: Determine the contribution of knowledge retrieval and reasoning adjustment.}

The accuracy of fast and slow thinking is evaluated to measure the retrieval capability $C_{\text{KR}}$ and reasoning adjustment capability $C_{\text{RA}}$:

\begin{equation}
C_{KR} \coloneqq A_{\text{fast}} = \frac{1}{|D|} \sum_{x\in D} \mathbb{I}(y_{\text{fast}}=y^*)
\label{definition of fast}
\end{equation}

\begin{equation}
C_{KR} + C_{RA} \coloneqq A_{\text{slow}} = \frac{1}{|D|} \sum_{x\in D} \mathbb{I}(y_{\text{slow}}=y^*)
\label{definition of slow}
\end{equation}

$A_{fast}$ and $A_{slow}$ denote the accuracy of fast and slow thinking on dataset $D$, $y^*$ denotes the true answer, and $|D|$ indicates the total number of problems. The indicator function $\mathbb{I} (\cdot)$ returns 1 if the condition is satisfied, and 0 otherwise. By subtracting Equation \ref{definition of fast} from Equation \ref{definition of slow}, we obtain:

\begin{equation}
C_{RA} \coloneqq \delta = A_{\text{slow}} - A_{\text{fast}} = \frac{1}{|D|} \sum_{x\in D} \mathbb{I}(y_{\text{slow}}=y^*) - \frac{1}{|D|} \sum_{x\in D} \mathbb{I}(y_{\text{fast}}=y^*)
\label{definition of reasoning}
\end{equation}

This equation shows that the accuracy difference between slow and fast thinking, referred to as the reasoning gain $\delta$, reflects the capability of reasoning adjustment $C_{RA}$. Ultimately, we decouple the knowledge and reasoning capabilities in Equation \ref{definition of fast} and Equation \ref{definition of reasoning}.

\textbf{Step 4: Decompose reasoning adjustment into correction and overthinking.}

Based on the properties of the indicator function:

\begin{equation}
\left\{
\begin{aligned}
    \sum_{x \in D} \mathbb{I}(y_{\text{slow}} = y^*) &= \sum_{x \in D} \mathbb{I}(y_{\text{slow}} = y^*) \cdot \left( \mathbb{I}(y_{\text{fast}} = y^*) + \mathbb{I}(y_{\text{fast}} \neq y^*) \right) \\
    \sum_{x \in D} \mathbb{I}(y_{\text{fast}} = y^*) &= \sum_{x \in D} \mathbb{I}(y_{\text{fast}} = y^*) \cdot \left( \mathbb{I}(y_{\text{slow}} = y^*) + \mathbb{I}(y_{\text{slow}} \neq y^*) \right)
\end{aligned}
\right.
\end{equation}

$\delta$ is derived as

\begin{equation}
\delta =  
\underbrace{\frac{1}{|D|} \sum_{x \in D} \mathbb{I}(y_{\text{fast}} \neq y^*) \cdot \mathbb{I}(y_{\text{slow}} = y^*)}_{\delta_{c}}
- 
\underbrace{\frac{1}{|D|} \sum_{x \in D} \mathbb{I}(y_{\text{fast}} = y^*) \cdot \mathbb{I}(y_{\text{slow}} \neq y^*)}_{\delta_{o}}
\label{equ:delta decomposition initial}
\end{equation}

This shows reasoning adjustment has two effects: it corrects the errors made by fast thinking, but also incorrectly override the correct answers. We refer to these effects as correction and overthinking, denoting their corresponding accuracy changes as $\delta_{c}$ and $\delta_{o}$. Their formal definitions are derived as:

\begin{equation}
\left\{
\begin{aligned}
\delta_{c} &= \frac{\sum_{x \in D} \mathbb{I}(y_{\text{fast}} \neq y^*) \cdot \mathbb{I}(y_{\text{slow}} = y^*)}{\sum_{x \in D} \mathbb{I}(y_{\text{fast}} \neq y^*)} \cdot \sum_{x \in D} \mathbb{I}(y_{\text{fast}} \neq y^*) = r_{c} \cdot |D_{fast}^{false}| \\
\delta_{o} &= \frac{\sum_{x \in D} \mathbb{I}(y_{\text{fast}} = y^*) \cdot \mathbb{I}(y_{\text{slow}} \neq y^*)}{\sum_{x \in D} \mathbb{I}(y_{\text{fast}} = y^*)} \cdot \sum_{x \in D} \mathbb{I}(y_{\text{fast}} = y^*)  = r_{o} \cdot |D_{fast}^{true}|
\end{aligned}
\right.
\end{equation}

$|D_{fast}^{false}|$ and $|D_{fast}^{true}|$ denote the numbers of incorrect and correct answers produced by fast thinking, respectively. $r_{c}$ and $r_{o}$ represent the rates at which the LLMs correct wrong answers and overthink correct answers. Thus, the reasoning gain $\delta$ is expressed as:

\begin{equation}
\delta = \delta_{c} - \delta{o} = 
\frac{1}{|D|} \Bigg( 
r_{c} \cdot |D_{fast}^{false}|
- 
r_{o} \cdot |D_{fast}^{true}|
\Bigg)
\label{equ:trade off}
\end{equation}

$\delta$ arises from two opposing components: the benefit from correction $\delta_{c}$ and the loss from overthinking $\delta_{o}$. $|D_{\text{fast}}^{\text{false}}|$ and $|D_{\text{fast}}^{\text{true}}|$, determined by knowledge retrieval, define the upper bounds for correction and overthinking. $\delta_{c}$ and $\delta_{o}$, influenced by reasoning adjustment, define the actual levels of correction and overthinking in the candidates.

Some research suggests reasoning benefits LLMs by providing information gain \cite{ton2024understanding}, while others attribute reasoning failure to noise introduction \cite{gan2025rethinking}. Equation \ref{equ:trade off} integrates these insights, indicating it is the trade off between information gain and noise that determines the effect of reasoning.

\section{Results}
\label{Results}

\subsection{Experiment Procedure}
\label{Experiment Procedure}
\paragraph{Dataset} MMLU, MathQA, and MedQA are employed \cite{hendrycks2020measuring,zhang2018medical,amini2019mathqa}. MMLU serves as a general dataset, while MathQA and MedQA are domain-specific datasets (Table \ref{tab:question_quantity}). Although each MMLU question is annotated with a subject, the 57-in-total subjects are too fine-grained for our analysis. Therefore, we group them into 13 broader domains, as shown in Table \ref{Domain classification and corresponding problem quantity}. For example, questions annotated with college physics, conceptual physics, or high school physics are grouped under the "Physics" domain.

\paragraph{Model}
To enhance the generalizability of our work, extensive experiments are conducted on 15 LLMs, including \textit{Qwen}, \textit{LLaMA}, \textit{Gemma}, \textit{Phi}, and \textit{GLM} \cite{yang2024qwen2,grattafiori2024llama,team2024gemma,abdin2024phi,glm2024chatglm}. Their abbreviations and full model names are shown in Appendix \ref{Model details}. Additional experimental details are provided in Appendix \ref{Experimental details}.

\subsection{Small LLMs Overthink More Than Correct During Reasoning}

Although LLMs have been extensively studied on MMLU, most existing studies fail to disentangle the contributions of knowledge retrieval and reasoning adjustment. Our cognition attribution architecture is employed to decouple their contribution on MMLU.

Table \ref{Model Performance} reports the accuracy of fast and slow thinking in "Fast" and "Slow" columns. Reasoning gain $\delta$ is determined by the accuracy difference between fast and slow thinking, and subsequently decomposed into $\delta_{c}$ and $\delta_{o}$ according to Equation \ref{equ:trade off}.

Reasoning adjustment benefits most models, yielding positive $\delta$. However, for some extremely small models (marked in red), $\delta$ becomes zero or even negative, as the loss from overthinking $\delta_{o}$ outweighs the benefit from correction $\delta_{c}$.

The correction rate $r_{c}$ and overthinking rate $r_{o}$ are investigated. Small LLMs exhibit lower $r_{c}$ and higher $r_{o}$ compared to their larger counterparts. Moreover, the variation in $r_{o}$ across model sizes is more prominent than that in $r_{o}$. Notably, the $r_{o}$ of \textit{LLaMA 1B} is 45.4\% higher than that of \textit{LLaMA 70B}, while its $r_{c}$ is merely 8.7\% lower. This asymmetry suggests that the substantial overthinking tendency, rather than the modest correction capability, plays a greater role in the negative reasoning gain in small models. More discussion on correction and overthinking is in Section \ref{Parameter Scaling Makes Reasoning Adjustment More Prudent than Smart}.

Table \ref{tab:token consumption} further exhibits that although small models consume a comparable number of tokens to their larger counterparts, they achieve limited reasoning gain $\delta$, suggesting that they acquire only the superficial form of reasoning without its substance.

Our results offer new insights into prior findings. \cite{liu2024mind, cuadron2025danger} report that reasoning negatively affects performance, while \cite{gan2025rethinking} attributes this to noise introduction. Our study further reveals that small LLMs are prone to introducing noise and leading to overthinking. \cite{li2025small} finds that small models benefit less from long CoT distillation. Our results suggest that is because small models introduce more noise when CoT becomes longer. In summary, our study indicates that small LLMs tend to introduce noise during reasoning, reflecting a lack of "prudence" during inference.

\begin{table}[!h]
\centering
\caption{Performance on MMLU using cognition attribution architecture. "Fast" and "Slow" columns represent the accuracy under fast and slow thinking, respectively. "$\delta$" column measures reasoning gain, decomposed into correction gain $\delta_{c}$ and overthinking loss $\delta_{o}$. "$r_{c}$" and "$r_{o}$" columns report correction rate and overthink rate, respectively.}
\label{Model Performance}
\begin{tabular}{lccccccc}
\toprule
\textbf{Model} & \textbf{Fast} & \textbf{Slow} & \textbf{$\delta$} & \textbf{$\delta_c$} & \textbf{$\delta_o$} & \textbf{$r_{c}$} & \textbf{$r_{o}$} \\
\midrule
\textit{Qwen 1.5B} & 53.9 & 50.0 & \textcolor{red}{-3.9} & 13.8 & 17.7 & 29.9 & 32.8 \\
\textit{Qwen 3B} & 60.9 & 63.0 & 2.1 & 14.1 & 12.0 & 35.9 & 19.7 \\
\textit{Qwen 7B} & 67.8 & 71.8 & 4.0 & 11.8 & 7.7 & 36.5 & 11.4 \\
\textit{Qwen 14B} & 75.2 & 78.7 & 3.6 & 9.1 & 5.5 & 36.5 & 7.3 \\
\textit{Qwen 32B} & 79.5 & 81.1 & 1.5 & 6.9 & 5.4 & 33.9 & 6.8 \\
\midrule
\textit{QwQ 32B} & 77.7 & 85.6 & 7.9 & 10.0 & 10.0 & 54.6 & 5.5 \\
\midrule
\textit{LLaMA 1B} & 35.1 & 35.1 & \textcolor{red}{0.1} & 17.9 & 17.3 & 29.4 & 52.6 \\
\textit{LLaMA 3B} & 52.4 & 55.9 & 3.5 & 17.1 & 13.6 & 35.9 & 25.9 \\
\textit{LLaMA 8B} & 60.3 & 65.7 & 5.4 & 15.9 & 10.4 & 40.0 & 17.3 \\
\textit{LLaMA 70B} & 81.1 & 82.4 & 1.3 & 7.2 & 5.9 & 38.1 & 7.2 \\
\midrule
\textit{Gemma 2B} & 53.5 & 50.9 & \textcolor{red}{-2.6} & 10.4 & 13.0 & 22.5 & 24.4 \\
\textit{Gemma 9B} & 69.3 & 69.5 & 0.3 & 9.5 & 9.2 & 30.8 & 13.3 \\
\textit{Gemma 27B} & 72.5 & 73.5 & 1.0 & 9.4 & 8.4 & 34.2 & 11.6 \\
\midrule
\textit{GLM 9B} & 63.8 & 70.1 & 6.3 & 14.5 & 8.2 & 40.0 & 13.0 \\
\midrule
\textit{Phi 14B} & 78.1 & 84.1 & 6.0 & 10.2 & 4.2 & 46.7 & 5.4 \\
\bottomrule
\end{tabular}
\end{table}

\subsection{Reasoning Exhibits Significantly Domain-Specific Variability}

While it is widely hypothesized that reasoning adjustment is domain-specific (e.g., mathematics is regarded as reasoning-intensive), the claim remains imprecise and speculative. We employ cognition attribution architecture to quantify the contribution of reasoning adjustment across domains.

Figure \ref{tab:domain_rankings} reports the reasoning gain $\delta$ for 15 LLMs across 13 domains, spanning natural sciences to humanities. The results reveal substantial cross-domain variation in $\delta$. Notably, \textit{Qwen 1.5B} shows an inter-domain accuracy gap of 34.7\% (22.3\% for Mathematics and -12.4\% for Political Science).

While domains exhibit varying sensitivity to reasoning adjustment, their relative ranking remains stable. The top-1, top-2, and top-3 $\delta$ for each model are highlighted in red, blue, and green, respectively. Across 15 models, $\delta$ of mathematics, physics, and chemistry ranks among the top-3 for 15, 14, and 12 times, respectively, indicating that these domains consistently benefit the most from reasoning adjustment. In contrast, political science and history yield negative $\delta$ in 10 and 9 models, respectively, suggesting that reasoning adjustment even impair their performance.

Our study provides quantitative evidence that reasoning adjustment exhibits significant domain-specific variability. This phenomenon is explained according to the relationship between correction and overthinking in Equation \ref{equ:trade off}. In knowledge-intensive domains, LLMs make mistakes due to insufficient knowledge. Reasoning cannot compensate for the knowledge gap and instead introduces additional noise. Consequently, $\delta_{c}$ falls short of $\delta_{o}$, yielding negative $\delta$ in these domains. In reasoning-intensive domains (notably mathematics, physics, and chemistry), mistakes often arise from generating answers under high uncertainty. Reasoning effectively reduce ambiguity and correct mistakes. Therefore, $\delta_{c}$ outweighs $\delta_{o}$, resulting in positive $\delta$ in these domains.

The results on MathQA and MedQA show high consistency with MMLU (Figure \ref{fig:mathqa}). A significantly high $\delta$ (with a range from 20.9\% to 40.2\%) is observed on MathQA, while the $\delta$ on MedQA ranges sorely from -8.5\% to 14.3\%.

\begin{table}[!h]
\setlength{\tabcolsep}{4pt}  
\centering
\caption{$\delta$ of 19 LLMs across 13 domains on MMLU. For each model, the top-1, top-2, and top-3 $\delta$ are highlighted in red, blue, and green, respectively. Mat, Mathematics; Phy, Physics; Che, Chemistry; CS, Computer Science; EcoBus, Economics and Business; Phi: Philosophy; GeoAst: Geography and Astronomy; BioMed, Biology and Medicine; PhySoc: Psychology and Sociology; Eng, Engineering; His: History; Pol: Political Science.}
\label{tab:domain_rankings}
\begin{tabular}{lccccccccccccc}
\toprule
\textbf{Model} & \textbf{Mat} & \textbf{Phy} & \textbf{Che} & \textbf{CS} & \textbf{\makecell{Eco\\Bus}} & \textbf{Phi} & \textbf{\makecell{Geo\\Ast}} & \textbf{\makecell{Bio\\Med}} & \textbf{\makecell{Psy\\Soc}} & \textbf{Eng} & \textbf{Law} & \textbf{His} & \textbf{Pol} \\
\midrule
\textit{Qwen 1.5B} & \textcolor{red}{22.3} & \textcolor{blue}{7.2} & \textcolor{green}{5.6} & -2.2 & -4.1 & -7.4 & -5.6 & -6.8 & -7.5 & -6.9 & -6.9 & -9.0 & -12.4 \\
\textit{Qwen 3B} & \textcolor{red}{25.8} & \textcolor{blue}{15.4} & \textcolor{green}{10.9} & 2.9 & 0.3 & 8.3 & 2.7 & -4.7 & -2.3 & -3.3 & 0.8 & -3.9 & -7.1 \\
\textit{Qwen 7B} & \textcolor{red}{27.6} & \textcolor{blue}{12.1} & \textcolor{green}{10.6} & 8.3 & 3.2 & 10.3 & -0.2 & -0.6 & -0.6 & -0.8 & 0.2 & -1.8 & -1.9 \\
\textit{Qwen 14B} & \textcolor{red}{20.0} & \textcolor{green}{12.3} & \textcolor{blue}{13.5} & 7.4 & 2.3 & 2.3 & 1.5 & 1.8 & 1.7 & 4.5 & 0.9 & -1.0 & -1.9 \\
\textit{Qwen 32B} & \textcolor{red}{13.9} & \textcolor{blue}{8.8} & \textcolor{green}{6.6} & 6.4 & 2.7 & 3.8 & -0.6 & -1.1 & -1.1 & -1.2 & -2.3 & -2.0 & 0.5 \\
\midrule
\textit{QwQ 32B} & \textcolor{red}{22.7} & \textcolor{green}{16.9} & 12.6 & 11.5 & 6.1 & \textcolor{blue}{19.7} & 2.3 & 4.6 & 3.4 & 6.7 & 10.5 & 4.1 & -1.5 \\
\midrule
\textit{LLaMA 1B} & \textcolor{green}{5.9} & \textcolor{red}{7.3} & -0.8 & -5.2 & 4.0 & -2.1 & 4.7 & 4.3 & \textcolor{blue}{6.3} & -0.2 & -7.7 & -2.0 & -3.8 \\
\textit{LLaMA 3B} & \textcolor{red}{20.6} & 7.6 & \textcolor{green}{8.3} & -1.6 & 2.4 & 0.6 & \textcolor{blue}{13.4} & 5.9 & 0.2 & -3.7 & -2.7 & 1.1 & 1.2 \\
\textit{LLaMA 8B} & \textcolor{red}{22.5} & \textcolor{blue}{15.0} & \textcolor{green}{10.6} & 5.8 & 4.7 & 1.9 & 4.9 & 4.0 & 4.9 & -0.4 & 0.6 & 0.7 & 2.8 \\
\textit{LLaMA 70B} & \textcolor{red}{16.4} & \textcolor{green}{7.0} & \textcolor{blue}{9.2} & 6.7 & 3.1 & -0.1 & -2.0 & -0.4 & -0.2 & -2.5 & -2.4 & -0.7 & -0.3 \\
\midrule
\textit{Gemma 2B} & \textcolor{blue}{4.9} & \textcolor{red}{5.7} & \textcolor{green}{-0.7} & -3.3 & -3.6 & -7.7 & -5.7 & -2.3 & -2.9 & -6.9 & -3.6 & -4.2 & -10.0 \\
\textit{Gemma 9B} & \textcolor{red}{12.7} & \textcolor{blue}{10.0} & \textcolor{green}{8.9} & 2.9 & -0.2 & -3.5 & -2.0 & -2.2 & -2.1 & -2.9 & -3.0 & -1.9 & -5.7 \\
\textit{Gemma 27B} & \textcolor{red}{19.9} & \textcolor{blue}{10.3} & 5.0 & \textcolor{green}{5.2} & 0.7 & -4.7 & -2.9 & -2.9 & -3.1 & -0.4 & -0.5 & -0.7 & -6.8 \\
\midrule
\textit{GLM 9B} & \textcolor{red}{24.5} & \textcolor{blue}{16.8} & \textcolor{green}{16.2} & 11.2 & 5.4 & 3.7 & 3.7 & 4.0 & 2.5 & 9.4 & -0.8 & -0.5 & 1.4 \\
\midrule
\textit{Phi 14B} & \textcolor{red}{25.9} & \textcolor{blue}{19.5} & \textcolor{green}{12.2} & 10.6 & 6.9 & 2.3 & 1.7 & 3.3 & 1.0 & 5.3 & 3.8 & 1.5 & 0.2 \\
\midrule
Average & 19.0 & 11.5 & 8.6 & 4.4 & 2.3 & 1.8 & 1.1 & 0.5 & 0.0 & -0.2 & -0.9 & -1.4 & -3.0 \\
\bottomrule
\end{tabular}
\end{table}

\subsection{Parameter Scaling Benefits Knowledge More than Reasoning}

\begin{figure}[!h]
    \centering
    \includegraphics[width=0.75\linewidth]{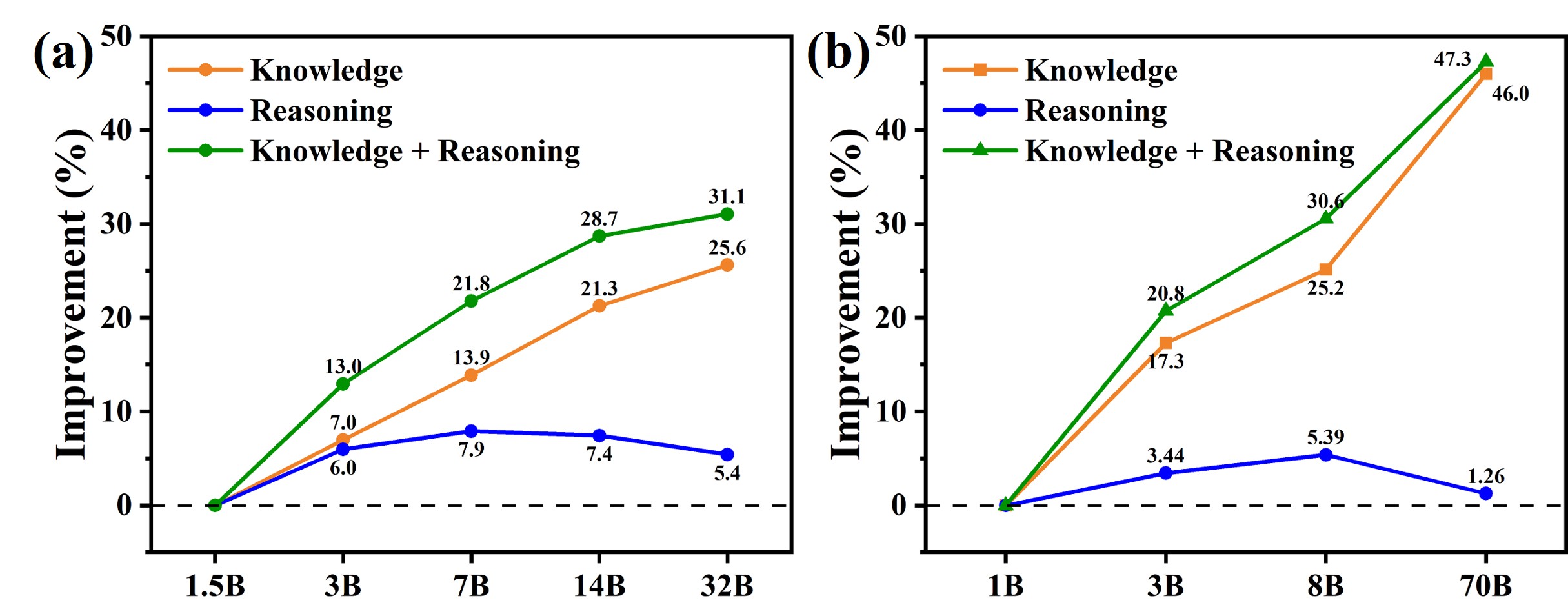}
    \caption{Improvement of knowledge retrieval and reasoning adjustment with parameter scaling. The plotted values quantify improvement relative to the smallest baseline model in each series: \textit{Qwen 1.5B} for the \textit{Qwen} series in panel (a) and \textit{LLaMA 1B} for the \textit{LLaMA} series in panel (b). The green, orange, and blue trajectories represent the $C_{KR} + C_{RA}$, $C_{KR}$, and $C_{RA}$ according to Equation \ref{definition of fast}-\ref{definition of reasoning}.}
    \label{fig:scaling}
\end{figure}

Although scaling law demonstrates parameter scaling improves LLMs' capabilities \cite{kaplan2020scaling}, it is still unclear scaling benefits knowledge or reasoning. To clarify this, the capability improvement from scaling is investigated in Figure \ref{fig:scaling}, with values representing the relative improvement compared to the smallest model in each series. Capabilities of knowledge and reasoning are determined according to Equation \ref{definition of fast}-\ref{definition of reasoning}. For instance, the fast thinking accuracy of \textit{Qwen 32B} and its smallest counterpart \textit{Qwen 1.5B} are 79.5\% and 53.9\%. The knowledge improvement from scaling becomes 25.6\%.

The results reveal that although both knowledge retrieval and reasoning adjustment benefit from model scaling, the improvement in knowledge retrieval is more pronounced and sustained. Reasoning adjustment shows notable gains primarily when scaling from small-sized models (1B) to medium-sized models (8B), but exhibits limited improvement beyond this range. Table \ref{Model Performance} demonstrates the observed enhancement in reasoning adjustment from small to medium size is largely attributed to a reduction in overthinking. Once the model reaches medium size, it becomes sufficiently "prudent" to avoid overthinking. Consequently, further scaling leads to only marginal reductions in overthinking, resulting in limited additional gains in reasoning adjustment. The next section further discusses the impact of scaling on correction and overthinking.

\subsection{Scaling Makes Reasoning Significantly More Prudent and Modestly More Intelligent}
\label{Parameter Scaling Makes Reasoning Adjustment More Prudent than Smart}

\begin{figure}[!h]
    \centering
    \includegraphics[width=0.75\linewidth]{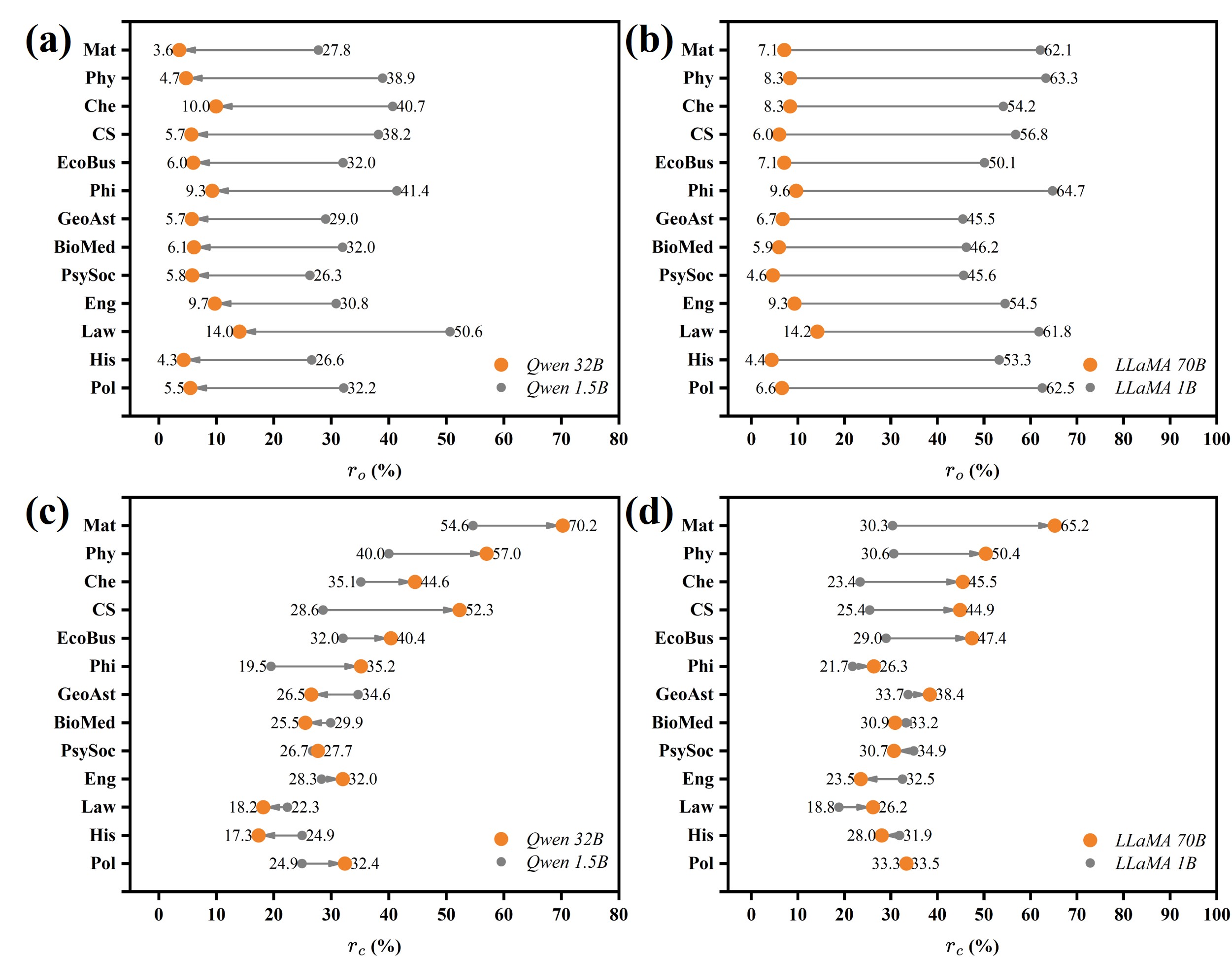}
    \caption{Impact of parameter scaling on overthinking rate $r_{o}$ and correction rate $r_{c}$. (a) and (b) show the $r_{o}$ of \textit{Qwen} and \textit{LLaMA}, respectively. (c) and (d) show the $r_{c}$ of \textit{Qwen} and \textit{LLaMA}, respectively. Mat, Mathematics; Phy, Physics; Che, Chemistry; CS, Computer Science; EcoBus, Economics and Business; Phi: Philosophy; GeoAst: Geography and Astronomy; BioMed, Biology and Medicine; PhySoc: Psychology and Sociology; Eng, Engineering; His: History; Pol: Political Science.}
    \label{fig:parameter scaling on overthinking and correction}
\end{figure}

Figure \ref{fig:parameter scaling on overthinking and correction} demonstrates the different scaling dynamics of overthinking and correction. In both (a) \textit{Qwen} and (b) \textit{LLaMA}, the overthinking rate $r_{o}$ significantly decreases across all domains, while (c) and (d) show that the correction rate $r_{c}$ only increases modestly in specific domains.

This phenomenon reveals an asymmetry in the effect of parameter scaling on different reasoning behaviors. The substantial reduction in overthinking rate across all domains suggests that scaling primarily enhances the model's prudence — specifically, its ability to avoid introducing noise. In particular, with larger models possessing greater knowledge capacity, the introduction of erroneous knowledge during COT reasoning is reduced, leading to less overthinking.

In contrast, the relatively modest and domain-specific improvement in correction rate implies that the correction capability of LLMs is less responsive to scaling compared to overthinking. This may reflect the fact that correction involves complex multi-step reasoning, which benefits less from parameter growth alone and require targeted training.

The anchor experiment in Appendix \ref{anchor experiment} further reveals that LLMs display human-like anchoring effects, being influenced by initial incorrect intuitions. Importantly, reasoning adjustment helps counteract these effects, making LLMs more prudent.

\subsection{Reasoning Occurs in Deeper Layers of LLMs}

The following reveals a phenomenon termed "cognitive hierarchy", indicating that knowledge retrieval and reasoning adjustment operate at different hierarchical layers within neural networks. To quantify this, we employ Centered Kernel Alignment (CKA), a measure of neural network similarity \cite{cortes2012algorithms,kornblith2019similarity}.

\begin{figure}[htbp]
    \centering
    \includegraphics[width=0.75\linewidth]{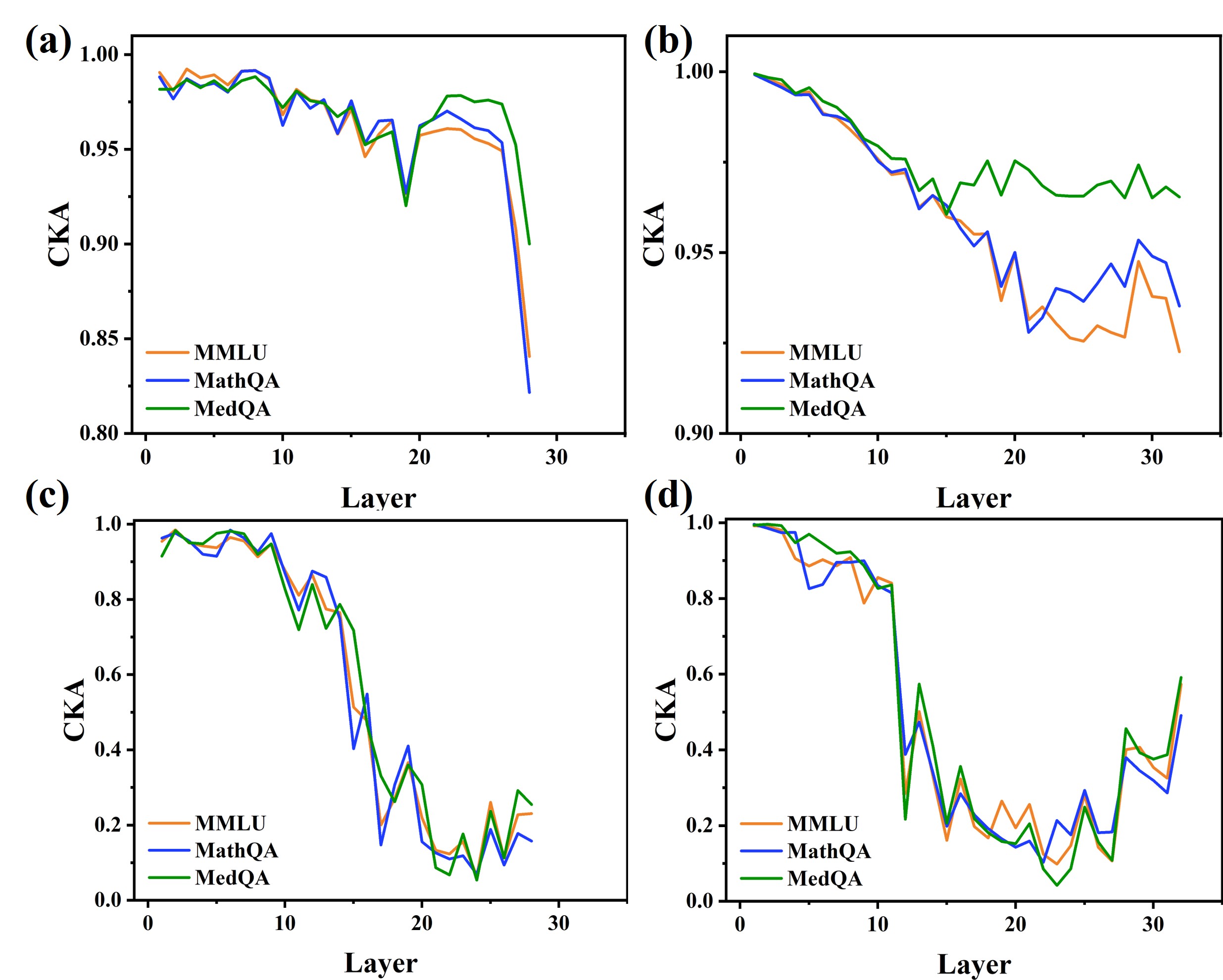}
    \caption{CKA between corresponding layers under slow and fast thinking. Lower CKA indicates reduced similarity for the given layer. (a, b) Output similarity for \textit{Qwen 7B} and \textit{LLaMA 8B}. (c, d) Attention similarity for \textit{Qwen 7B} and \textit{LLaMA 8B}.}
    \label{fig:cka}
\end{figure}

Figure \ref{fig:cka} presents the CKA between the same neural network layer under slow and fast thinking, where a lower CKA indicates reduced similarity for the corresponding layer. (a) and (b) exhibit output CKA for \textit{Qwen 7B} and \textit{LLaMA 8B}, while (c) and (d) exhibits attention CKA. More details about CKA are in Appendix \ref{Details of CKA}.

All CKA curves exhibit an initial plateau in lower layers, followed by a decline in higher layers, indicating that lower layers remain similar across thinking modes, while higher layers diverge. This pattern suggests that both modes share knowledge retrieval in lower layers, but slow thinking additionally engages reasoning adjustment in higher layers. Thus, our results suggest knowledge retrieval and reasoning adjustment are primarily localized to lower and higher layers, respectively.

These findings align with recent studies on knowledge editing \cite{zhang2024co,meng2022locating,meng2022mass}, which identify knowledge as being primarily located in the lower layers. Our study extends this by demonstrating that reasoning adjustment occurs in the higher layers.

\section{Related Works}
\label{gen_inst}

\paragraph{CoT} To enhance the reasoning capability of LLMs, CoT and its variants, including Tree-of-Thought (ToT), Program-of-Thought (PoT), and Graph-of-Thought, have been proposed \cite{wei2022chain,yao2023tree,chen2022program,besta2024graph}, which endow LLMs with human-like cognitive abilities. Early approaches rely on prompt engineering to explicitly guide CoT generation, but these methods suffer from poor generalizability and heavy dependence on domain expertise. Although techniques like Zero-Shot CoT and Auto-CoT are proposed \cite{kojima2022large,zhang2022automatic}, it remains challenging to endow LLMs with the capability to automatically generate high-quality CoT.

\paragraph{Reasoning LLMs} The emergence of OpenAI's O1 demonstrates the feasibility of large models autonomously generating high-quality Chain-of-Thoughts (CoTs) \cite{openai2023learningtoreason}. While the implementation details of O1 remain undisclosed, several researchers have conducted extensive investigations into its capabilities \cite{qin2024o1,huang2024o1,huang2025o1}. Initially, researchers propose utilizing external models to supervise and guide the generation of CoT \cite{cobbe2021training}. The most representative methods combine Monte Carlo Tree Search (MCTS) with process-supervised reward models \cite{zhang2024rest,guan2025rstar}. Although these methods effectively enhance LLMs' performance on complex reasoning tasks, MCTS suffers from prohibitive computational latency \cite{jiang2024technical}, and training process-supervised models requires massive CoT datasets with step-by-step annotations, which are extremely costly to obtain \cite{setlur2024rewarding,lu2024autopsv,wang2023math}. Subsequent efforts aim to internalize reasoning capabilities within LLMs themselves. Studies find that distilling just 3.9K CoT examples elevates a non-reasoning model's performance to match that of reasoning-specialized models \cite{min2024imitate}. Recently, DeepSeek shows that outcome-based rewards suffice to enable LLMs to generate high-quality CoT autonomously, significantly reducing the training cost of reasoning-capable LLMs \cite{guo2025deepseek,qwen2025qwq32b,team2025kimi}. Consequently,LLMs possess not only extensive knowledge but also advanced reasoning abilities.

\section{Conclusion}
\label{conclusion}

We present a cognition attribution architecture that disentangles knowledge and reasoning in LLMs. In this framework, the cognition of LLMs is decomposed into two phase according to dual-system cognitive theory: knowledge retrieval (Phase 1) and reasoning adjustment (Phase 2). Subsequently, LLMs are prompted to generate answers under different cognitive modes, including fast thinking and slow thinking. The performance under fast thinking determine the capability of knowledge retrieval, while the performance gain attributed to slow thinking quantifies the capability of reasoning adjustment. Ultimately, reasoning adjustment is decomposed into correction and overthinking.

Using this architecture, our findings reveal: (1) reasoning adjustment is domain-specific, benefiting reasoning-intensive domains but potentially harming knowledge-intensive domains; (2) parameter scaling primarily enhances both knowledge retrieval and reasoning adjustment, with reasoning adjustment more significant. Moreover, parameter scaling make reasoning adjustment significantly more prudent in all domains and modestly more intelligent in specific domains. (3) knowledge retrieval and reasoning adjustment are hierarchically organized, residing in lower and higher network layers, respectively. Our architecture not only offers new a perspective on the cognitive properties of LLMs, but also provides insights into some existing research, such as scaling law, hierarchical knowledge editing, and reasoning limitations of small model.

\paragraph{Limitations} First, our evaluation is limited to multiple-choice QA; future work should extend the analysis to open-ended generation. Second, while we study 15 models, ultra-large models (>100B) are not included. Third, although we define and quantify knowledge and reasoning based on the dual-system theory, our methods are still simplified and may not fully capture the underlying mechanisms in LLMs. Moreover, this study primarily demonstrates the application of our framework. Although we have observed some interesting phenomena using the framework, due to space limitations, a more in-depth exploration is not feasible at this stage, which remains a direction for future work.

\section{Acknowledgments}
The work is supported by Noncommunicable Chronic Diseases-National Science and Technology Major Project (Grant No. 2023ZD0506501). 

\newpage

\bibliography{neurips_2025}
\bibliographystyle{ieeetr}

\appendix

\section{Prompt}
\label{prompt}
\subsection{Fast thinking prompt}
Fast thinking requires LLMs to directly output answers based on its knowledge without engaging in any additional CoT. The prompt that guides LLMs to conduct
fast thinking prompt is:

\begin{figure}[!h]
    \centering
    \includegraphics[width=0.75\linewidth]{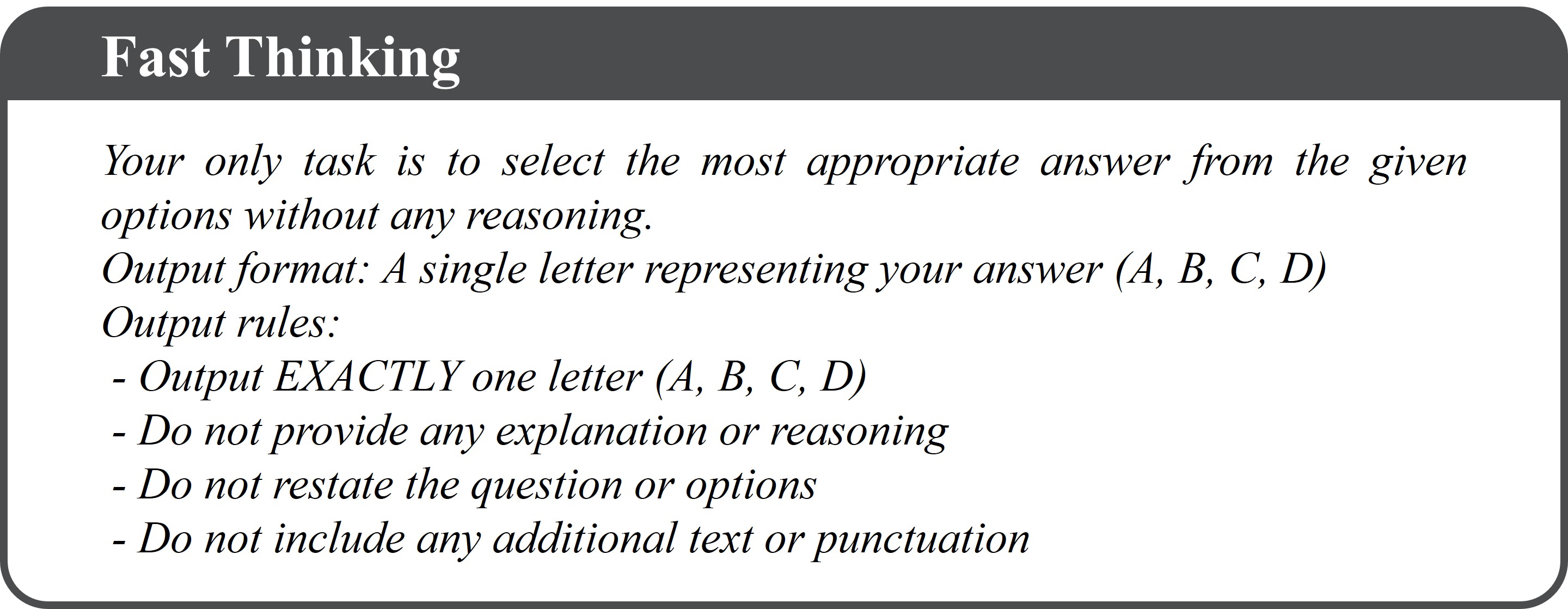}
    \caption{Fast thinking prompt}
    \label{fig:enter-label}
\end{figure}

\subsection{Fast thinking example}
\label{fast thinking exmaple}
Question: A university’s mathematics department has 10 professors and will offer 20 different courses next semester. Each professor will be assigned to teach exactly 2 of the courses, and each course will have exactly one professor assigned to teach it. If any professor can be assigned to teach any course, how many different complete assignments of the 10 professors to the 20 courses are possible?

Option:

A: 20!/2\textasciicircum(10)

B: 10!/2\textasciicircum(9)

C: 10\textasciicircum(20) - 2\textasciicircum(10)

D: 10\textasciicircum(20) - 100

Fast thinking answer:

A

\subsection{Slow thinking prompt}

Slow thinking requires LLMs to adjust the intuitive answer generated by fast thinking after reasoning. For reasoning LLMs, including Qwen QwQ Preview and DeepSeek-Distilled Qwen, the question is directly input without any additional prompt guidance. For other non-reasoning LLMs, following prompt is used to guide slow thinking:

\begin{figure}[!h]
    \centering
    \includegraphics[width=0.75\linewidth]{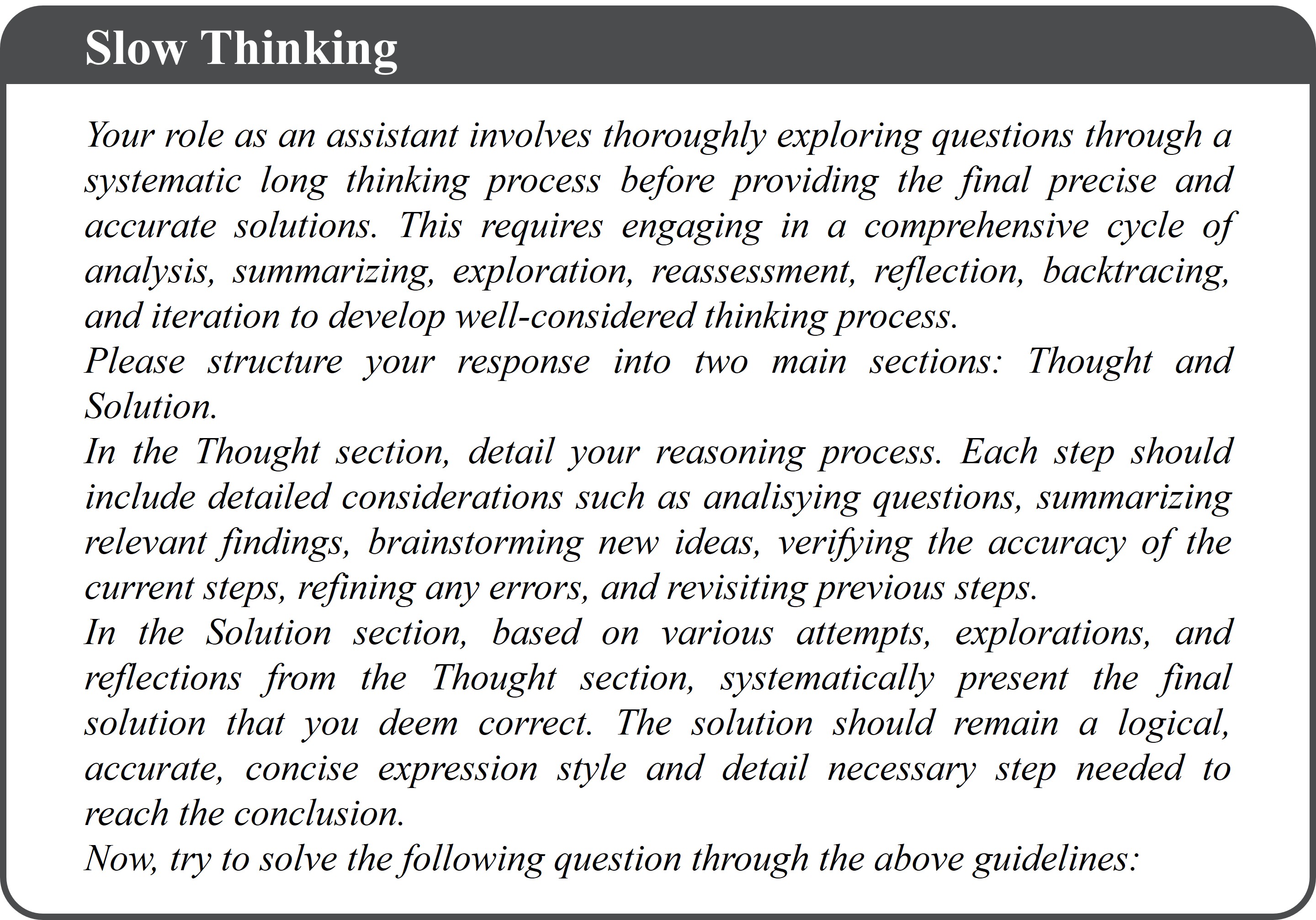}
    \caption{Slow thinking prompt}
    \label{fig:enter-label}
\end{figure}

\subsection{Slow thinking example}
\label{slow thinking exmaple}

Question: A university’s mathematics department has 10 professors and will offer 20 different courses next semester. Each professor will be assigned to teach exactly 2 of the courses, and each course will have exactly one professor assigned to teach it. If any professor can be assigned to teach any course, how many different complete assignments of the 10 professors to the 20 courses are possible?

Option:

A: 20!/2\textasciicircum(10)

B: 10!/2\textasciicircum(9)

C: 10\textasciicircum(20) - 2\textasciicircum(10)

D: 10\textasciicircum(20) - 100

Slow thinking answer:

\#\#\# Thought

1. **Analyzing the Question:**

- The problem involves assigning 10 professors to 20 courses, with each professor teaching exactly 2 courses and each course being taught by exactly one professor.

- The key is to understand the combinatorial nature of the problem, which involves selecting and assigning courses to professors.

2. **Summarizing Relevant Findings:**

- Each professor is assigned 2 courses out of 20, and there are 10 professors.

- The total number of courses is 20, and each course is assigned to exactly one professor.

- The problem can be approached by considering the number of ways to choose 2 courses for each professor and then arranging the professors to these courses.

3. **Brainstorming New Ideas:**

- One approach is to consider the number of ways to choose 2 courses out of 20 for the first professor, then 2 out of the remaining 18 for the second professor, and so on.

- Another approach is to consider the total permutations of assigning 20 courses to 10 professors, but this needs to be adjusted for the fact that each professor is assigned exactly 2 courses.

4. **Verifying the Accuracy of Current Steps:**

- The first approach involves calculating combinations for each professor and then multiplying these combinations.

- The second approach involves calculating permutations and then dividing by the number of ways to arrange the 2 courses for each professor (since the order of courses for each professor does not matter).

5. **Refining Any Errors:**

- The first approach might overcount because it does not account for the indistinguishability of the order of courses for each professor.

- The second approach is more accurate as it accounts for the indistinguishability of the order of courses for each professor.

6. **Revisiting Previous Steps:**

- The correct approach is to calculate the total permutations of assigning 20 courses to 10 professors and then divide by the number of ways to arrange the 2 courses for each professor (2! for each professor).

\#\#\# Solution

1. **Total Permutations of Assigning Courses:**

- The total number of ways to assign 20 courses to 10 professors is (20!).

2. **Adjusting for Indistinguishability:**

- Since each professor is assigned exactly 2 courses, and the order of these 2 courses does not matter, we need to divide by (2!) for each professor.

- There are 10 professors, so we divide by (2\textasciicircum(10)).

3. **Final Calculation:**

- The number of different complete assignments is frac\{20!\}\{2\textasciicircum(10)\}.

Thus, the correct answer is:**A: 20!/2\textasciicircum(10)**

\section{Data}
\label{Data}

This study involves three datasets, including MMLU, MathQA and MedQA. Only the test part of MMLU is involved. MMLU serves as general dataset, while MathQA and MedQA represent the domain-specific datasets.

\begin{table}[!h]
\centering
\setlength{\tabcolsep}{10pt} 
\caption{Question quantity of dataset}
\label{tab:question_quantity}
\begin{tabular}{lc}
\hline
\textbf{Dataset} & \textbf{Question Quantity} \\ \hline
MMLU  & 14042 \\
MathQA & 2985 \\
MedQA  & 1273 \\ \hline
\end{tabular}
\end{table}

MMLU comprises 57 distinct topics, grouped into 13 fields based on disciplinary similarity. Among them, Miscellaneous is not assigned to any specific field. The topic of human aging primarily addresses demographic trends rather than biological mechanisms and is therefore categorized under Psychology and Sociology instead of Biology and Medicine. Computer Security focuses on security architecture rather than algorithms, and is thus classified separately from Computer Science.

\begin{longtable}[htbp]{llc}
\caption{Domain classification and corresponding problem quantity}
\label{Domain classification and corresponding problem quantity} \\
\hline
\textbf{Field} & \textbf{Subject} & \textbf{Problem Quantity} \\
\hline
\endfirsthead

\hline
\textbf{Field} & \textbf{Subject} & \textbf{Problem Quantity} \\
\hline
\endhead

\hline
\endfoot

\hline
\endlastfoot

\multirow{5}{*}{Mathematics} 
    & Abstract Algebra & 100 \\
    & College Mathematics & 100 \\
    & Elementary Mathematics & 378 \\
    & High School Mathematics & 270 \\
    & High School Statistics & 216 \\

\hline 

\multirow{3}{*}{Physics}
    & College Physics & 102 \\
    & Conceptual Physics & 235 \\
    & High School Physics & 151 \\

\hline 

\multirow{2}{*}{Chemistry}
    & College Chemistry & 100 \\
    & High School Chemistry & 203 \\

\hline 

\multirow{3}{*}{Computer Science}
    & College Computer Science & 100 \\
    & High School Computer Science & 100 \\
    & Machine Learning & 112 \\

\hline 

\multirow{7}{*}{Economics and Business}
    & Business Ethics & 100 \\
    & Econometrics & 114 \\
    & High School Macroeconomics & 390 \\
    & High School Microeconomics & 238 \\
    & Management & 103 \\
    & Marketing & 234 \\
    & Professional Accounting & 282 \\

\hline 

\multirow{10}{*}{Biology and Medicine}
    & Anatomy & 135 \\
    & Clinical Knowledge & 265 \\
    & College Biology & 144 \\
    & College Medicine & 173 \\
    & High School Biology & 310 \\
    & Human Sexuality & 131 \\
    & Medical Genetics & 100 \\
    & Nutrition & 306 \\
    & Professional Medicine & 272 \\
    & Virology & 166 \\

\hline 

\multirow{4}{*}{Psychology and Sociology}
    & High School Psychology & 545 \\
    & Human Aging & 223 \\
    & Professional Psychology & 612 \\
    & Sociology & 201 \\

\hline 

\multirow{4}{*}{Geography and Astronomy}
    & Astronomy & 152 \\
    & Global Facts & 100 \\
    & High School Geography & 198 \\
    & World Religions & 171 \\

\hline 

\multirow{2}{*}{Engineering}
    & Computer Security & 100 \\
    & Electrical Engineering & 145 \\

\hline 

\multirow{5}{*}{Philosophy}
    & Formal Logic & 126 \\
    & Logical Fallacies & 163 \\
    & Moral Disputes & 346 \\
    & Moral Scenarios & 895 \\
    & Philosophy & 311 \\

\hline 

\multirow{3}{*}{Law}
    & International Law & 121 \\
    & Jurisprudence & 108 \\
    & Professional Law & 1534 \\

\hline 

\multirow{4}{*}{History}
    & High School European History & 165 \\
    & High School US History & 204 \\
    & High School World History & 237 \\
    & Prehistory & 324 \\

\hline 

\multirow{4}{*}{Political Science}
    & High School Government and Politics & 193 \\
    & Public Relations & 110 \\
    & Security Studies & 245 \\
    & US Foreign Policy & 100 \\

\end{longtable}

\section{Model details}
\label{Model details}

The abbreviations of LLMs are as follows:

\begin{table}[!h]
\centering
\setlength{\tabcolsep}{10pt} 
\caption{Model name abbreviations}
\label{tab:model_names}
\begin{tabular}{ll}
\hline
\textbf{Full Name} & \textbf{Abbreviation} \\ \hline
\textit{Qwen 2.5} & \textit{Qwen} \\
\textit{Qwen QwQ 32B Preview} & \textit{QwQ 32B} \\
\textit{LlaMA 3} & \textit{LLaMA} \\
\textit{Gemma 2} & \textit{Gemma} \\
\textit{GLM 4} & \textit{GLM} \\
\textit{Phi 4} & \textit{Phi} \\ \hline
\end{tabular}
\end{table}

\section{Experimental details}
\label{Experimental details}
\subsection{Hyperparamters}
To minimize uncertainty, the temperature of the LLM is set to 0, with Pass@1 accuracy serving as the final performance metric for the model. The experiments are conducted using Python 3.10.13, vLLM 0.6.6, and Transformers 4.47.1.

\subsection{Evaluation methods}

When evaluating the correctness of fast thinking, the model's responses are directly compared against the ground-truth answers. When assessing slow thinking, GLM-4-PLUS is employed as the evaluator. The prompt used for evaluation is as follows:

\begin{figure}[!h]
    \centering
    \includegraphics[width=0.75\linewidth]{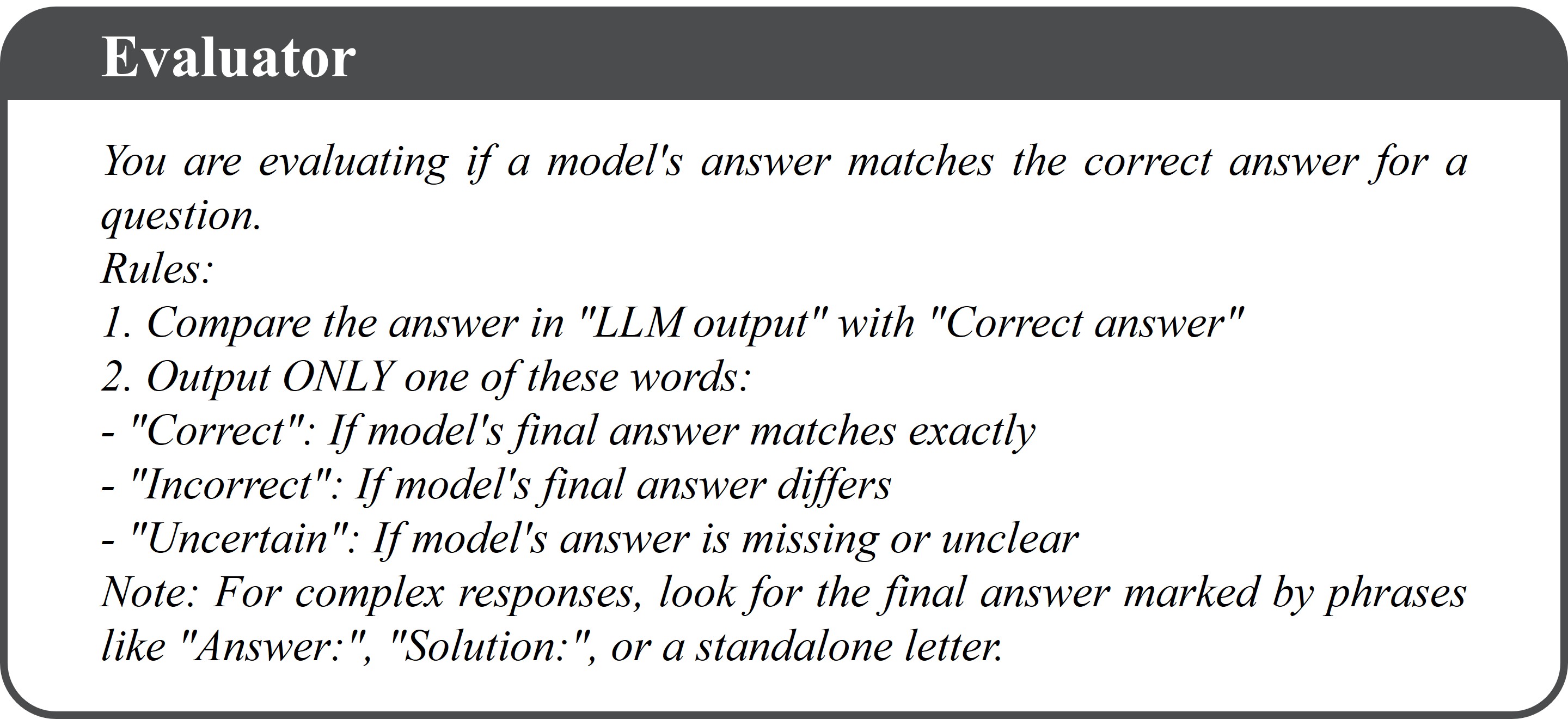}
    \caption{Evaluation prompt}
    \label{fig:enter-label}
\end{figure}

\section{Results of Token consumption}
\label{Results of Token consumption}

Table \ref{tab:token consumption} presents the token consumption of various LLMs on MMLU. The "Correct", "Incorrect", and "Overall" columns represent token consumption for correct answers, incorrect answers, and both, respectively. The results show that although smaller models consume a similar number of tokens as larger models, they achieve limited reasoning gain ($\delta$). Furthermore, incorrect answers generally lead to higher token consumption than correct answers.\\
\\
\\
\begin{table}[!h]
\setlength{\tabcolsep}{20pt} 
\caption{Token consumption of LLMs on MMLU. The "Correct", "Incorrect", and "Overall" columns indicate the token consumption corresponding to the correct answers, incorrect answers, and both correct and incorrect answers, respectively.}
\label{tab:token consumption}
\centering
\begin{tabular}{lccc}
\hline
\textbf{Model} & \textbf{Correct} & \textbf{Incorrect} & \textbf{Overall} \\
\hline
\textit{Qwen 1.5B} & 473 & 546 & 510 \\
\textit{Qwen 3B} & 520 & 616 & 556 \\
\textit{Qwen 7B} & 552 & 607 & 568 \\
\textit{Qwen 14B} & 565 & 630 & 579 \\
\textit{Qwen 32B} & 515 & 557 & 523 \\
\hline
\textit{QwQ 32B} & 1050 & 1538 & 1120 \\
\hline
\textit{LLaMA 1B} & 540 & 686 & 635 \\
\textit{LLaMA 3B} & 569 & 759 & 653 \\
\textit{LLaMA 8B} & 595 & 869 & 689 \\
\textit{LLaMA 70B} & 503 & 664 & 532 \\
\hline
\textit{Gemma 2B} & 602 & 674 & 637 \\
\textit{Gemma 9B} & 452 & 523 & 474 \\
\textit{Gemma 27B} & 422 & 488 & 440 \\
\hline
\textit{GLM 9B} & 536 & 609 & 558 \\
\hline
\textit{Phi 14B} & 586 & 693 & 603 \\
\hline
\end{tabular}
\end{table}

\section{Detailed results across domains}
\subsection{Detailed results on MMLU}

\begin{longtable}{l*{2}{c}*{2}{c}*{2}{c}*{2}{c}*{2}{c}}
\caption{The accuracy of the \textit{Qwen} series across various domains}
\label{tab:qwen_accuracy} \\
\toprule
\multirow{2}{*}{} & \multicolumn{2}{c}{\textit{Qwen 3B}} & \multicolumn{2}{c}{\textit{Qwen 7B}} & \multicolumn{2}{c}{\textit{Qwen 14B}} & \multicolumn{2}{c}{\textit{Qwen 32B}} & \multicolumn{2}{c}{\textit{QwQ 32B}} \\
\cmidrule(lr){2-3} \cmidrule(lr){4-5} \cmidrule(lr){6-7} \cmidrule(lr){8-9} \cmidrule(l){10-11}
 & Fast & Slow & Fast & Slow & Fast & Slow & Fast & Slow & Fast & Slow \\
\midrule
\endfirsthead

\toprule
\multirow{2}{*}{} & \multicolumn{2}{c}{\textit{Qwen 3B}} & \multicolumn{2}{c}{\textit{Qwen 7B}} & \multicolumn{2}{c}{\textit{Qwen 14B}} & \multicolumn{2}{c}{\textit{Qwen 32B}} & \multicolumn{2}{c}{\textit{QwQ 32B}} \\
\cmidrule(lr){2-3} \cmidrule(lr){4-5} \cmidrule(lr){6-7} \cmidrule(lr){8-9} \cmidrule(l){10-11}
 & Fast & Slow & Fast & Slow & Fast & Slow & Fast & Slow & Fast & Slow \\
\midrule
\endhead

\bottomrule
\endfoot

Abstract algebra & 30 & 29 & 41 & 54 & 51 & 66 & 58 & 73 & 61 & 83 \\
College mathematics & 38 & 43 & 42 & 54 & 42 & 71 & 53 & 80 & 67 & 83 \\
Elementary mathematics & 43 & 77 & 52 & 88 & 63 & 94 & 79 & 96 & 88 & 96 \\
High school mathematics & 34 & 62 & 44 & 79 & 47 & 85 & 56 & 87 & 66 & 90 \\
High school statistics & 45 & 57 & 60 & 69 & 69 & 81 & 73 & 83 & 80 & 86 \\
\hline
College physics & 40 & 45 & 57 & 69 & 51 & 73 & 57 & 82 & 70 & 86 \\
Conceptual physics & 48 & 56 & 58 & 72 & 73 & 79 & 79 & 85 & 86 & 90 \\
High school physics & 33 & 40 & 42 & 62 & 59 & 74 & 70 & 83 & 72 & 82 \\
\hline
College chemistry & 34 & 38 & 48 & 47 & 43 & 56 & 53 & 61 & 56 & 63 \\
High school chemistry & 41 & 48 & 49 & 66 & 60 & 69 & 67 & 83 & 76 & 83 \\
\hline
College computer science & 43 & 40 & 56 & 45 & 58 & 68 & 67 & 78 & 78 & 87 \\
High school computer science & 60 & 53 & 70 & 78 & 82 & 89 & 87 & 91 & 90 & 94 \\
Machine learning & 37 & 39 & 38 & 48 & 53 & 61 & 61 & 68 & 71 & 77 \\
\hline
Business ethics & 62 & 51 & 66 & 64 & 73 & 69 & 76 & 75 & 80 & 78 \\
Econometrics & 31 & 39 & 54 & 41 & 58 & 57 & 66 & 63 & 68 & 65 \\
High school macroeconomics & 53 & 55 & 63 & 67 & 73 & 78 & 81 & 84 & 85 & 87 \\
High school microeconomics & 59 & 48 & 72 & 72 & 83 & 83 & 86 & 87 & 91 & 92 \\
Management & 72 & 67 & 75 & 71 & 83 & 84 & 84 & 83 & 81 & 85 \\
Marketing & 82 & 71 & 84 & 79 & 90 & 87 & 90 & 88 & 91 & 91 \\
Professional accounting & 41 & 36 & 49 & 56 & 54 & 67 & 64 & 73 & 65 & 76 \\
\hline
Anatomy & 52 & 50 & 64 & 53 & 73 & 69 & 70 & 76 & 77 & 73 \\
Clinical knowledge & 61 & 56 & 64 & 67 & 75 & 76 & 82 & 81 & 86 & 83 \\
College biology & 62 & 62 & 74 & 70 & 81 & 78 & 88 & 88 & 94 & 94 \\
College medicine & 62 & 55 & 68 & 63 & 71 & 72 & 77 & 83 & 81 & 83 \\
High school biology & 67 & 64 & 81 & 76 & 84 & 84 & 89 & 87 & 93 & 91 \\
Human sexuality & 60 & 55 & 71 & 63 & 76 & 77 & 82 & 84 & 87 & 86 \\
Medical genetics & 70 & 50 & 74 & 74 & 83 & 77 & 85 & 90 & 91 & 92 \\
Nutrition & 60 & 52 & 69 & 68 & 76 & 74 & 81 & 85 & 83 & 83 \\
Professional medicine & 53 & 43 & 71 & 60 & 75 & 76 & 84 & 84 & 88 & 88 \\
Virology & 45 & 36 & 51 & 45 & 50 & 52 & 52 & 52 & 55 & 50 \\
\hline
High school psychology & 77 & 70 & 84 & 81 & 86 & 88 & 89 & 92 & 92 & 92 \\
Human aging & 57 & 52 & 70 & 61 & 70 & 69 & 75 & 77 & 78 & 76 \\
Professional psychology & 53 & 46 & 60 & 61 & 74 & 70 & 78 & 80 & 82 & 80 \\
Sociology & 74 & 62 & 79 & 74 & 82 & 84 & 88 & 85 & 88 & 88 \\
\hline
Astronomy & 67 & 57 & 70 & 71 & 85 & 81 & 85 & 89 & 89 & 91 \\
Global facts & 23 & 33 & 33 & 46 & 41 & 53 & 52 & 56 & 57 & 51 \\
High school geography & 70 & 57 & 73 & 77 & 85 & 83 & 90 & 89 & 90 & 89 \\
World religions & 76 & 73 & 81 & 78 & 83 & 81 & 85 & 86 & 89 & 89 \\
\hline
Computer security & 71 & 59 & 67 & 66 & 74 & 73 & 77 & 82 & 84 & 79 \\
Electrical engineering & 52 & 48 & 57 & 52 & 66 & 65 & 70 & 74 & 77 & 78 \\
\hline
Formal logic & 48 & 41 & 42 & 44 & 59 & 56 & 61 & 67 & 70 & 73 \\
Logical fallacies & 74 & 56 & 76 & 64 & 83 & 80 & 88 & 85 & 86 & 86 \\
Moral disputes & 54 & 50 & 59 & 59 & 70 & 70 & 74 & 75 & 81 & 78 \\
Moral scenarios & 31 & 22 & 28 & 42 & 33 & 48 & 65 & 68 & 68 & 73 \\
Philosophy & 56 & 54 & 66 & 59 & 70 & 71 & 79 & 78 & 78 & 81 \\
\hline
International law & 64 & 60 & 75 & 69 & 76 & 74 & 79 & 81 & 83 & 85 \\
Jurisprudence & 69 & 57 & 72 & 69 & 76 & 76 & 81 & 83 & 87 & 84 \\
Professional law & 36 & 29 & 40 & 42 & 48 & 49 & 54 & 54 & 60 & 58 \\
\hline
High school european history & 72 & 66 & 76 & 73 & 83 & 77 & 84 & 85 & 84 & 83 \\
High school us history & 65 & 57 & 80 & 75 & 84 & 84 & 89 & 87 & 92 & 91 \\
High school world history & 72 & 58 & 80 & 70 & 84 & 80 & 89 & 86 & 92 & 89 \\
Prehistory & 60 & 52 & 69 & 69 & 75 & 75 & 84 & 85 & 88 & 86 \\
\hline
High school government and politics & 73 & 63 & 79 & 74 & 91 & 91 & 93 & 92 & 94 & 95 \\
Public relations & 58 & 45 & 63 & 59 & 66 & 62 & 74 & 72 & 73 & 72 \\
Security studies & 59 & 44 & 68 & 58 & 70 & 68 & 77 & 74 & 78 & 79 \\
Us foreign policy & 75 & 64 & 76 & 67 & 82 & 79 & 90 & 89 & 93 & 92 \\
\end{longtable}

\begin{longtable}{l*{2}{c}*{2}{c}*{2}{c}*{2}{c}}
\caption{The accuracy of the \textit{LLaMA} series across various domains}
\label{tab:llama_accuracy} \\ 
\toprule
\multirow{2}{*}{} & \multicolumn{2}{c}{\textit{LLaMA 1B}} & \multicolumn{2}{c}{\textit{LLaMA 3B}} & \multicolumn{2}{c}{\textit{LLaMA 8B}} & \multicolumn{2}{c}{\textit{LLaMA 70B}} \\
\cmidrule(lr){2-3} \cmidrule(lr){4-5} \cmidrule(lr){6-7} \cmidrule(l){8-9}
 & Fast & Slow & Fast & Slow & Fast & Slow & Fast & Slow \\
\midrule
\endfirsthead

\toprule
\multirow{2}{*}{} & \multicolumn{2}{c}{\textit{LLaMA 1B}} & \multicolumn{2}{c}{\textit{LLaMA 3B}} & \multicolumn{2}{c}{\textit{LLaMA 8B}} & \multicolumn{2}{c}{\textit{LLaMA 70B}} \\
\cmidrule(lr){2-3} \cmidrule(lr){4-5} \cmidrule(lr){6-7} \cmidrule(l){8-9}
 & Fast & Slow & Fast & Slow & Fast & Slow & Fast & Slow \\
\midrule
\endhead

\bottomrule
\endfoot

\bottomrule
\endfoot

Abstract algebra & 27 & 17 & 30 & 36 & 33 & 35 & 61 & 71 \\
College mathematics & 30 & 17 & 38 & 27 & 31 & 40 & 53 & 67 \\
Elementary mathematics & 28 & 49 & 41 & 77 & 47 & 84 & 78 & 93 \\
High school mathematics & 22 & 27 & 34 & 61 & 40 & 65 & 56 & 83 \\
High school statistics & 24 & 20 & 40 & 49 & 48 & 56 & 73 & 83 \\
\hline
College physics & 22 & 19 & 32 & 44 & 36 & 59 & 64 & 82 \\
Conceptual physics & 29 & 40 & 46 & 55 & 53 & 70 & 82 & 83 \\
High school physics & 20 & 28 & 36 & 40 & 43 & 50 & 68 & 77 \\
\hline
College chemistry & 23 & 27 & 36 & 41 & 41 & 50 & 57 & 65 \\
High school chemistry & 33 & 30 & 45 & 55 & 48 & 59 & 72 & 82 \\
\hline
College computer science & 36 & 27 & 52 & 44 & 57 & 53 & 69 & 77 \\
High school computer science & 40 & 36 & 56 & 58 & 65 & 72 & 91 & 95 \\
Machine learning & 29 & 29 & 34 & 35 & 38 & 51 & 66 & 74 \\
\hline
Business ethics & 37 & 33 & 57 & 53 & 65 & 67 & 83 & 76 \\
Econometrics & 29 & 28 & 35 & 34 & 42 & 45 & 65 & 61 \\
High school macroeconomics & 25 & 36 & 50 & 53 & 59 & 68 & 84 & 88 \\
High school microeconomics & 31 & 33 & 49 & 61 & 66 & 69 & 90 & 92 \\
Management & 32 & 45 & 59 & 69 & 74 & 77 & 87 & 88 \\
Marketing & 43 & 52 & 73 & 78 & 81 & 87 & 93 & 93 \\
Professional accounting & 27 & 23 & 44 & 37 & 51 & 52 & 64 & 77 \\
\hline
Anatomy & 45 & 47 & 56 & 58 & 64 & 70 & 81 & 80 \\
Clinical knowledge & 37 & 43 & 55 & 64 & 67 & 72 & 84 & 84 \\
College biology & 35 & 39 & 55 & 67 & 68 & 76 & 92 & 92 \\
College medicine & 33 & 43 & 55 & 63 & 65 & 68 & 80 & 80 \\
High school biology & 37 & 44 & 56 & 68 & 71 & 80 & 92 & 93 \\
Human sexuality & 38 & 40 & 63 & 51 & 70 & 73 & 85 & 85 \\
Medical genetics & 36 & 37 & 63 & 73 & 63 & 78 & 94 & 93 \\
Nutrition & 37 & 39 & 57 & 68 & 70 & 71 & 89 & 85 \\
Professional medicine & 37 & 42 & 67 & 68 & 76 & 76 & 92 & 92 \\
Virology & 31 & 30 & 49 & 42 & 49 & 48 & 52 & 54 \\
\hline
High school psychology & 36 & 53 & 74 & 75 & 81 & 86 & 93 & 93 \\
Human aging & 32 & 35 & 60 & 59 & 65 & 66 & 83 & 81 \\
Professional psychology & 33 & 35 & 53 & 52 & 61 & 67 & 83 & 83 \\
Sociology & 44 & 40 & 68 & 69 & 73 & 78 & 90 & 90 \\
\hline
Astronomy & 41 & 41 & 53 & 64 & 72 & 68 & 93 & 89 \\
Global facts & 32 & 30 & 28 & 33 & 40 & 34 & 60 & 62 \\
High school geography & 36 & 44 & 56 & 71 & 68 & 79 & 90 & 89 \\
World religions & 51 & 47 & 72 & 74 & 81 & 83 & 91 & 92 \\
\hline
Computer security & 43 & 43 & 62 & 60 & 69 & 67 & 83 & 77 \\
Electrical engineering & 33 & 33 & 52 & 47 & 55 & 56 & 77 & 77 \\
\hline
Formal logic & 35 & 23 & 40 & 40 & 44 & 45 & 63 & 67 \\
Logical fallacies & 45 & 28 & 56 & 61 & 74 & 66 & 86 & 80 \\
Moral disputes & 36 & 37 & 50 & 51 & 63 & 64 & 83 & 80 \\
Moral scenarios & 30 & 19 & 32 & 30 & 35 & 43 & 88 & 81 \\
Philosophy & 36 & 35 & 56 & 57 & 63 & 66 & 78 & 79 \\
\hline
International law & 50 & 47 & 69 & 64 & 70 & 77 & 89 & 88 \\
Jurisprudence & 29 & 35 & 53 & 67 & 68 & 72 & 84 & 86 \\
Professional law & 32 & 23 & 42 & 38 & 49 & 48 & 68 & 66 \\
\hline
High school european history & 35 & 34 & 66 & 71 & 69 & 77 & 83 & 85 \\
High school us history & 44 & 42 & 63 & 66 & 77 & 77 & 94 & 90 \\
High school world history & 40 & 46 & 72 & 71 & 80 & 79 & 90 & 90 \\
Prehistory & 40 & 31 & 58 & 57 & 69 & 67 & 86 & 86 \\
\hline
High school government and politics & 34 & 44 & 65 & 74 & 78 & 82 & 96 & 95 \\
Public relations & 28 & 41 & 54 & 51 & 62 & 61 & 75 & 73 \\
Security studies & 41 & 27 & 53 & 52 & 59 & 67 & 76 & 78 \\
Us foreign policy & 53 & 32 & 72 & 67 & 85 & 77 & 91 & 91 \\
\end{longtable}

\begin{longtable}{l*{2}{c}*{2}{c}*{2}{c}}
\caption{The accuracy of \textit{Gemma} series across various domains}
\label{tab:gemma_accuracy} \\
\toprule
\multirow{2}{*}{} & \multicolumn{2}{c}{\textit{Gemma 2B}} & \multicolumn{2}{c}{\textit{Gemma 9B}} & \multicolumn{2}{c}{\textit{Gemma 27B}} \\
\cmidrule(lr){2-3} \cmidrule(lr){4-5} \cmidrule(l){6-7}
 & Fast & Slow & Fast & Slow & Fast & Slow \\
\midrule
\endfirsthead

\toprule
\multirow{2}{*}{} & \multicolumn{2}{c}{\textit{Gemma 2B}} & \multicolumn{2}{c}{\textit{Gemma 9B}} & \multicolumn{2}{c}{\textit{Gemma 27B}} \\
\cmidrule(lr){2-3} \cmidrule(lr){4-5} \cmidrule(l){6-7}
 & Fast & Slow & Fast & Slow & Fast & Slow \\
\midrule
\endhead

\bottomrule
\endfoot

Abstract algebra & 36 & 32 & 46 & 43 & 44 & 51 \\
College mathematics & 26 & 28 & 34 & 41 & 43 & 49 \\
Elementary mathematics & 42 & 61 & 59 & 83 & 57 & 89 \\
High school mathematics & 33 & 32 & 48 & 58 & 44 & 68 \\
High school statistics & 49 & 41 & 61 & 66 & 68 & 75 \\
\hline
College physics & 33 & 39 & 49 & 72 & 52 & 75 \\
Conceptual physics & 46 & 52 & 67 & 76 & 77 & 80 \\
High school physics & 31 & 36 & 58 & 61 & 57 & 70 \\
\hline
College chemistry & 43 & 44 & 54 & 54 & 49 & 59 \\
High school chemistry & 43 & 42 & 59 & 72 & 68 & 71 \\
\hline
College computer science & 39 & 33 & 49 & 56 & 63 & 64 \\
High school computer science & 53 & 58 & 82 & 79 & 77 & 86 \\
Machine learning & 36 & 28 & 47 & 52 & 65 & 71 \\
\hline
Business ethics & 54 & 47 & 67 & 66 & 72 & 71 \\
Econometrics & 36 & 33 & 61 & 54 & 60 & 49 \\
High school macroeconomics & 57 & 56 & 78 & 79 & 78 & 80 \\
High school microeconomics & 57 & 54 & 82 & 83 & 86 & 85 \\
Management & 74 & 65 & 82 & 80 & 83 & 81 \\
Marketing & 78 & 75 & 89 & 88 & 92 & 88 \\
Professional accounting & 39 & 34 & 56 & 57 & 57 & 68 \\
\hline
Anatomy & 51 & 50 & 67 & 70 & 75 & 75 \\
Clinical knowledge & 56 & 51 & 76 & 71 & 81 & 75 \\
College biology & 68 & 65 & 86 & 85 & 93 & 86 \\
College medicine & 60 & 59 & 75 & 73 & 78 & 76 \\
High school biology & 68 & 69 & 89 & 87 & 91 & 90 \\
Human sexuality & 61 & 55 & 83 & 73 & 82 & 76 \\
Medical genetics & 68 & 68 & 79 & 83 & 83 & 86 \\
Nutrition & 58 & 56 & 72 & 74 & 81 & 77 \\
Professional medicine & 54 & 51 & 78 & 75 & 83 & 82 \\
Virology & 47 & 42 & 52 & 45 & 54 & 46 \\
\hline
High school psychology & 80 & 76 & 91 & 90 & 92 & 92 \\
Human aging & 51 & 57 & 74 & 70 & 79 & 72 \\
Professional psychology & 51 & 48 & 76 & 74 & 81 & 76 \\
Sociology & 71 & 62 & 88 & 83 & 88 & 85 \\
\hline
Astronomy & 57 & 52 & 78 & 79 & 84 & 80 \\
Global facts & 26 & 27 & 36 & 42 & 48 & 47 \\
High school geography & 74 & 67 & 87 & 83 & 88 & 86 \\
World religions & 65 & 67 & 85 & 81 & 87 & 84 \\
\hline
Computer security & 66 & 54 & 80 & 69 & 76 & 74 \\
Electrical engineering & 51 & 48 & 63 & 66 & 67 & 68 \\
\hline
Formal logic & 40 & 33 & 51 & 48 & 56 & 52 \\
Logical fallacies & 66 & 57 & 84 & 81 & 85 & 78 \\
Moral disputes & 60 & 49 & 73 & 66 & 77 & 73 \\
Moral scenarios & 27 & 26 & 34 & 40 & 43 & 49 \\
Philosophy & 56 & 53 & 71 & 71 & 75 & 69 \\
\hline
International law & 63 & 48 & 79 & 74 & 81 & 79 \\
Jurisprudence & 62 & 57 & 80 & 81 & 82 & 77 \\
Professional law & 38 & 35 & 54 & 51 & 56 & 56 \\
\hline
High school european history & 67 & 68 & 81 & 81 & 82 & 78 \\
High school us history & 70 & 66 & 86 & 81 & 86 & 85 \\
High school world history & 70 & 65 & 89 & 82 & 87 & 89 \\
Prehistory & 59 & 53 & 76 & 78 & 80 & 79 \\
\hline
High school government and politics & 76 & 79 & 93 & 96 & 96 & 93 \\
Public relations & 56 & 52 & 71 & 65 & 68 & 61 \\
Security studies & 59 & 36 & 74 & 63 & 76 & 64 \\
Us foreign policy & 69 & 60 & 88 & 80 & 88 & 86 \\
\end{longtable}

\begin{longtable}{l*{2}{c}*{2}{c}}
\caption{The accuracy of \textit{GLM} and \textit{Phi} models across various domains}
\label{tab:glm_phi_accuracy} \\
\toprule
\multirow{2}{*}{} & \multicolumn{2}{c}{\textit{GLM 9B}} & \multicolumn{2}{c}{\textit{Phi 14B}} \\
\cmidrule(lr){2-3} \cmidrule(l){4-5}
 & Fast & Slow & Fast & Slow \\
\midrule
\endfirsthead

\toprule
\multirow{2}{*}{} & \multicolumn{2}{c}{\textit{GLM 9B}} & \multicolumn{2}{c}{\textit{Phi 14B}} \\
\cmidrule(lr){2-3} \cmidrule(l){4-5}
 & Fast & Slow & Fast & Slow \\
\midrule
\endhead

\bottomrule
\endfoot

Abstract algebra & 29 & 47 & 59 & 85 \\
College mathematics & 60 & 66 & 77 & 85 \\
Elementary mathematics & 70 & 75 & 87 & 93 \\
High school mathematics & 76 & 71 & 83 & 77 \\
High school statistics & 71 & 75 & 82 & 83 \\
\hline
College physics & 76 & 77 & 92 & 94 \\
Conceptual physics & 41 & 57 & 61 & 72 \\
High school physics & 47 & 64 & 68 & 88 \\
\hline
College chemistry & 35 & 40 & 53 & 80 \\
High school chemistry & 66 & 71 & 82 & 85 \\
\hline
College computer science & 36 & 69 & 54 & 96 \\
High school computer science & 78 & 84 & 79 & 87 \\
Machine learning & 61 & 73 & 82 & 92 \\
\hline
Business ethics & 44 & 41 & 65 & 69 \\
Econometrics & 53 & 65 & 79 & 82 \\
High school macroeconomics & 51 & 84 & 71 & 97 \\
High school microeconomics & 43 & 56 & 65 & 85 \\
Management & 41 & 40 & 50 & 61 \\
Marketing & 82 & 86 & 92 & 95 \\
Professional accounting & 57 & 73 & 75 & 88 \\
\hline
Anatomy & 78 & 81 & 89 & 92 \\
Clinical knowledge & 85 & 82 & 83 & 84 \\
College biology & 83 & 85 & 91 & 89 \\
College medicine & 88 & 91 & 95 & 97 \\
High school biology & 71 & 79 & 84 & 92 \\
Human sexuality & 34 & 65 & 54 & 93 \\
Medical genetics & 79 & 87 & 92 & 96 \\
Nutrition & 51 & 66 & 64 & 83 \\
Professional medicine & 86 & 90 & 93 & 94 \\
Virology & 56 & 70 & 77 & 86 \\
\hline
High school psychology & 86 & 84 & 89 & 92 \\
Human aging & 87 & 85 & 90 & 90 \\
Professional psychology & 72 & 67 & 78 & 77 \\
Sociology & 73 & 69 & 82 & 85 \\
\hline
Astronomy & 74 & 79 & 88 & 88 \\
Global facts & 74 & 73 & 90 & 87 \\
High school geography & 79 & 75 & 86 & 85 \\
World religions & 46 & 59 & 66 & 75 \\
\hline
Computer security & 79 & 83 & 85 & 86 \\
Electrical engineering & 89 & 88 & 91 & 92 \\
\hline
Formal logic & 70 & 68 & 90 & 93 \\
Logical fallacies & 82 & 84 & 88 & 94 \\
Moral disputes & 66 & 69 & 81 & 81 \\
Moral scenarios & 38 & 59 & 68 & 73 \\
Philosophy & 67 & 77 & 82 & 86 \\
\hline
International law & 68 & 69 & 85 & 83 \\
Jurisprudence & 69 & 71 & 85 & 87 \\
Professional law & 46 & 57 & 66 & 87 \\
\hline
High school european history & 47 & 46 & 62 & 66 \\
High school us history & 68 & 73 & 87 & 93 \\
High school world history & 65 & 69 & 84 & 85 \\
Prehistory & 64 & 63 & 76 & 74 \\
\hline
High school government and politics & 63 & 65 & 78 & 78 \\
Public relations & 82 & 84 & 89 & 89 \\
Security studies & 79 & 78 & 89 & 90 \\
Us foreign policy & 48 & 48 & 54 & 52 \\
\end{longtable}

\subsection{Detailed results on MathQA and MedQA}

In addition to MMLU, this paper also conducts experiments on MathQA and MedQA. To ensure comparability between datasets, the four-option version of MedQA is selected. For MathQA, one wrong option is randomly removed to reduce the choices from five to four.

The difference between slow and fast thinking in MathQA is more significant than that in MedQA, indicating reasoning contributes more on mathematics than medicine. This result is aligned with that on MMLU.

\begin{figure}[htbp]
    \centering
    \includegraphics[width=1\linewidth]{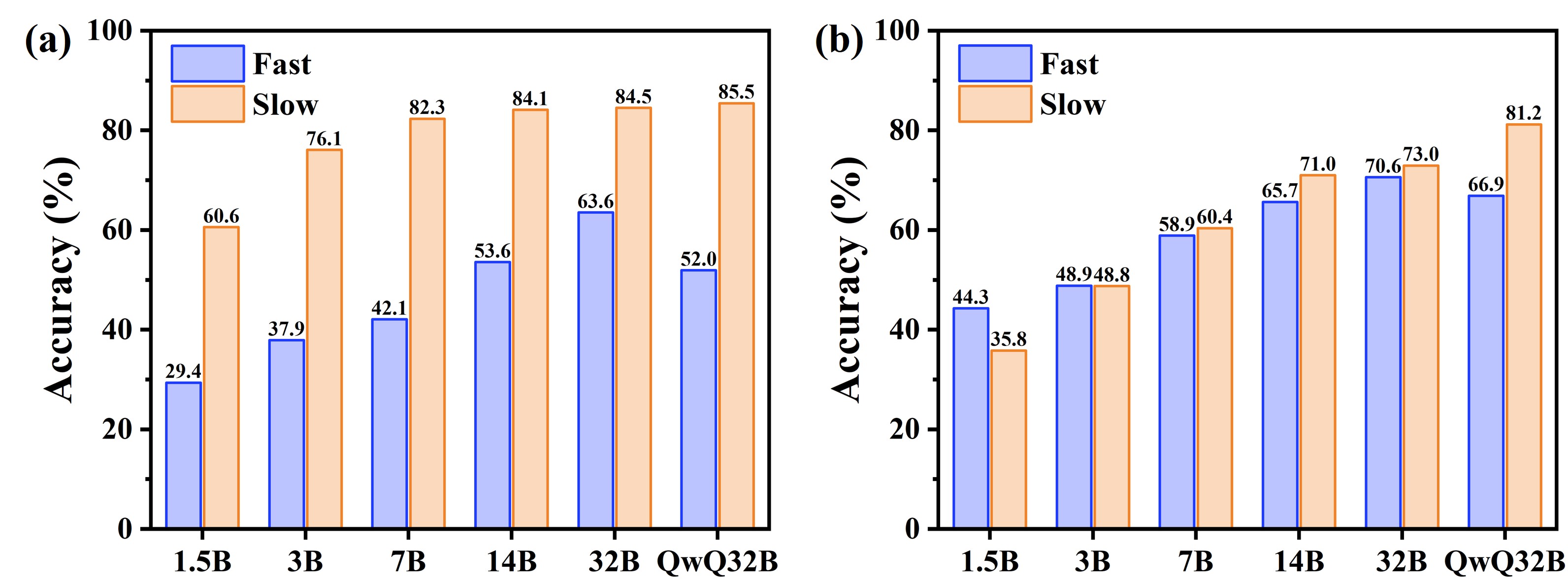}
    \caption{The accuracy of fast and slow thinking on (a) MathQA and (b) MedQA. The results demonstrate that mathematics imposes higher demands on reasoning than medicine.}
    \label{fig:mathqa}
\end{figure}

\section{Anchoring effect}
\label{anchor experiment}

\subsection{Methods}

According to cognitive science, humans are often influenced by initial information, a phenomenon known as the anchoring effect. To investigate whether LLMs exhibit similar behavior, a random incorrect option is added at the end of the problem. Specifically, for the following question:

Question: A university’s mathematics department has 10 professors and will offer 20 different courses next semester. Each professor will be assigned to teach exactly 2 courses, and each course will have one professor assigned to it. If any professor can be assigned to any course, how many different complete assignments of the 10 professors to the 20 courses are possible?

Option:

A: 20!/2\textasciicircum(10)

B: 10!/2\textasciicircum(9)

C: 10\textasciicircum(20) - 2\textasciicircum(10)

D: 10\textasciicircum(20) - 100

The correct answer is A. Therefore, a random incorrect answer, selected from options B, C, or D, is added at the end of the question. Assuming the randomly selected incorrect answer is B, the question becomes:

Question: A university’s mathematics department has 10 professors and will offer 20 different courses next semester. Each professor will be assigned to teach exactly 2 of the courses, and each course will have exactly one professor assigned to teach it. If any professor can be assigned to teach any course, how many different complete assignments of the 10 professors to the 20 courses are possible?

Option:

A: 20!/2\textasciicircum(10)

B: 10!/2\textasciicircum(9)

C: 10\textasciicircum(20) - 2\textasciicircum(10)

D: 10\textasciicircum(20) - 100

\textbf{The correct answer seems to be B.}

\subsection{Results}

It is important to emphasize that, as demonstrated earlier, reasoning plays a crucial role in the field of mathematics. Therefore, this study focuses solely on mathematics problems from MMLU.

Table \ref{anckor} shows that, after anchoring, LLM accuracy generally declines (marked in blue in Table \ref{anckor}), indicating that LLMs exhibit an anchoring effect similar to humans. Notably, the 1.5B model, when using fast thinking alone, achieved only 8\% accuracy, representing a 31\% drop. This suggests that fast thinking, which relies solely on knowledge retrieval, is highly susceptible to anchoring disruption.

In contrast, the performance decline in slow thinking is generally less severe, indicating that reasoning adjustment effectively mitigates the anchoring effect. For the 7B model, reasoning reduces a performance drop of 32\%.

\begin{table}[h!]
\centering
\caption{The "Anchor" and "w/o Anchor" rows represent the LLMs with and without anchoring, respectively. The "Anchor - w/o Anchor" row indicates the accuracy decrease due to anchoring.}
\label{anckor}
\begin{tabular}{llccc}
\hline
\textbf{Model Size} & \textbf{Process} & \textbf{Fast} & \textbf{Slow} & \textbf{$\delta$} \\
\hline
1.5B & Anchor & 8 & 40 & 32 \\
& w/o Anchor & 39 & 62 & 23 \\
& Anchor - w/o Anchor & \textcolor{blue}{-31} & \textcolor{blue}{-22} & +9 \\
\hline
3B & Anchor & 5 & 62 & 57 \\
& w/o Anchor & 50 & 76 & 26 \\
& Anchor - w/o Anchor & \textcolor{blue}{-45} & \textcolor{blue}{-14} & +31 \\
\hline
7B & Anchor & 26 & 78 & 52 \\
& w/o Anchor & 57 & 85 & 28 \\
& Anchor - w/o Anchor & \textcolor{blue}{-31} & \textcolor{blue}{-7} & +24 \\
\hline
14B & Anchor & 48 & 83 & 35 \\
& w/o Anchor & 67 & 87 & 20 \\
& Anchor - w/o Anchor & \textcolor{blue}{-19} & \textcolor{blue}{-4} & +15 \\
\hline
32B & Anchor & 50 & 83 & 33 \\
& w/o Anchor & 76 & 90 & 14 \\
& Anchor - w/o Anchor & \textcolor{blue}{-26} & \textcolor{blue}{-7} & +19 \\
\hline
\end{tabular}
\end{table}

\section{Details of CKA}
\label{Details of CKA}

To measure the similarity between fast and slow thinking, the CKA is calculated for each specific layer between the two modes. To ensure comparability, CKA is computed only on the tokens corresponding to the question part of the prompt, excluding those from the instruction part.

CKA is a metric used to compare the similarity between two sets of activations, typically derived from different neural network models or different layers of the same model. CKA leverages kernel methods to evaluate the alignment between the learned representations of the data. The calculation of CKA consists of several key steps:

\subsection*{1. Centering the Matrices}
The first step in CKA is centering the activation matrices. Given two matrices \(X\) and \(Y\) representing the activations of two models or layers, the centering operation involves subtracting the mean of each column from the respective elements. This centers the data and removes the mean shift, allowing for a more robust similarity measure. The centered matrices are denoted as \( \tilde{X} \) and \( \tilde{Y} \), where:

\begin{equation}
\tilde{X} = X - \frac{1}{n} \mathbf{1} \mathbf{1}^T X, \quad \tilde{Y} = Y - \frac{1}{n} \mathbf{1} \mathbf{1}^T Y
\end{equation}
where \( n \) is the number of samples, and \( \mathbf{1} \) is a vector of ones.

\subsection*{2. Gram Matrix Computation}
After centering, the next step is to compute the Gram matrices for both datasets. The Gram matrix represents the inner product between the centered data points. For matrices \( \tilde{X} \) and \( \tilde{Y} \), the Gram matrices are computed as:

\begin{equation}
K_X = \tilde{X} \tilde{X}^T, \quad K_Y = \tilde{Y} \tilde{Y}^T
\end{equation}

These Gram matrices capture the pairwise similarities between the data points in each representation.

\subsection*{3. Frobenius Norm}
The Frobenius norm of a matrix \(A\) is defined as the square root of the sum of the squared elements of \(A\):

\begin{equation}
\|A\|_F = \sqrt{\sum_{i,j} A_{ij}^2}
\end{equation}

For the Gram matrices \( K_X \) and \( K_Y \), we compute their Frobenius norms \( \|K_X\|_F \) and \( \|K_Y\|_F \), which are used to scale the alignment measure.

\subsection*{4. Hilbert-Schmidt Independence Criterion (HSIC)}
The HSIC is a measure of dependence between two datasets. It is computed by taking the inner product between the two centered Gram matrices:

\begin{equation}
\text{HSIC}(X, Y) = \sum_{i,j} K_{X,ij} K_{Y,ij}
\end{equation}

where \(K_{X,ij}\) and \(K_{Y,ij}\) are the elements of the Gram matrices \( K_X \) and \( K_Y \), respectively. HSIC quantifies the relationship between the activations of two representations.

\subsection*{5. CKA Calculation}
Finally, the CKA value is computed by normalizing the HSIC value with the product of the Frobenius norms of the Gram matrices. The formula for the CKA is:

\begin{equation}
\text{CKA}(X, Y) = \frac{\text{HSIC}(X, Y)}{\|K_X\|_F \cdot \|K_Y\|_F}
\end{equation}

The CKA score lies in the range \([0, 1]\), where a value closer to 1 indicates that the two representations are highly similar, and a value closer to 0 indicates dissimilarity.

\end{document}